%% file: main_camera_ready_v2.tex
\documentclass{article} % For LaTeX2e
\usepackage{iclr2026_conference,times}

\input{math_commands.tex}

\usepackage{hyperref}
\usepackage{subcaption}
\usepackage{url}
\usepackage{graphicx}
\usepackage{multirow}
\usepackage{booktabs}
\usepackage{amssymb}
\usepackage{algorithm}
\usepackage{algpseudocode}
\usepackage[table,xcdraw]{xcolor}

\definecolor{layer1}{HTML}{FEFEDA}
\definecolor{layer2}{HTML}{EDFEED}
\definecolor{layer3}{HTML}{EFF6FE}

\title{\textit{Mixture-of-Visual-Thoughts}: Exploring Context-Adaptive Reasoning Mode Selection for \\ General Visual Reasoning}
\author{Zejun Li$^{1,2,3\thefootnote{\dag}\ddag}$, Yingxiu Zhao$^{2,3\thefootnote{\dag}}$, Jiwen Zhang$^{1}$, Siyuan Wang$^{4}$, Yang Yao$^{2,3}$ \\ \textbf{Runzhou Zhao$^{2,3}$, Jun Song$^{2,3\thefootnote{*}}$, Bo Zheng$^{2,3}$, Zhongyu Wei$^{1,5\thefootnote{*}}$} \\
$^1$Fudan University \quad $^2$Alibaba Group Holding Limited \quad $^3$Future Living Lab of Alibaba \\
$^4$University of Southern California \qquad $^5$Shanghai Innovation Institute \\
\texttt{\{zejunli20,zywei\}@fudan.edu.cn \qquad jsong.sj@alibaba-inc.com} \\
\url{https://github.com/Future-Living-Lab/mixture-of-visual-thoughts}
}

\iclrfinalcopy % Uncomment for camera-ready version, but NOT for submission.
\begin{document}

\maketitle

\def\thefootnote{$\dag$}\footnotetext{Equal contribution; $^{\ddag}$ Work done during the internship at Future Living Lab of Alibaba.}

\def\thefootnote{*}\footnotetext{Corresponding authors.}

\begin{abstract}
Current visual reasoning methods mainly focus on exploring specific reasoning modes. Although improvements can be achieved in particular domains, they struggle to develop general reasoning capabilities. 
% Inspired by this, we propose a novel adaptive visual reasoning paradigm that unifies different reasoning modes within a single model and guides it to select the appropriate mode based on context. 
Inspired by this, we propose a novel adaptive reasoning paradigm, \underline{M}ixture-\underline{o}f-\underline{V}isual-\underline{T}houghts (\textbf{MoVT}), which unifies different reasoning modes within a single model and guides it to select the appropriate mode based on the context.
To achieve this, we introduce \textbf{AdaVaR}, a two-stage \underline{Ada}ptive \underline{V}isu\underline{a}l \underline{R}easoning learning framework: different modes are unified and learned during the supervised cold-start stage, and the mode selection capability is induced via an RL process with a carefully designed AdaGRPO algorithm.
Extensive experiments show that AdaVaR effectively guides the model to learn and differentiate multiple modes and perform context-adaptive mode selection, achieving consistent improvement across various scenarios, 
highlighting MoVT as an effective solution for building general visual reasoning models.
% highlighting adaptive reasoning as a key inductive bias for general-purpose visual reasoning.
\end{abstract}

\section{Introduction}
\label{section:intro}
% In recent years, significant progress has been achieved in enhancing the reasoning capabilities of large language models (LLMs). Researchers reveal the potential of autoregressive LLMs in performing multi-step reasoning and analysis by generating chains of thought (CoT) in various forms before producing final answers, thereby improving their ability to tackle complex problems~\citep{kojima2022large,wei2022chain,wang2023selfconsistency,yao2024tree}.
Recent research has revealed the reasoning potential of large language models (LLMs): by guiding LLMs to generate chain-of-thought (CoT) rationales in various forms before providing an answer, their performance on complex problems can be substantially improved~\citep{kojima2022large,wei2022chain,wang2023selfconsistency,yao2024tree}.
Furthermore, the reasoning capability of LLMs can be further enhanced through reinforcement learning (RL), ultimately pushing the frontier of model capability~\citep{guo2025deepseek,team2025kimi,yu2025dapoopensourcellmreinforcement}.

\begin{figure}[t]
  \centering
  \begin{subfigure}{0.47\linewidth}
  \centering
    \includegraphics[width=\linewidth]{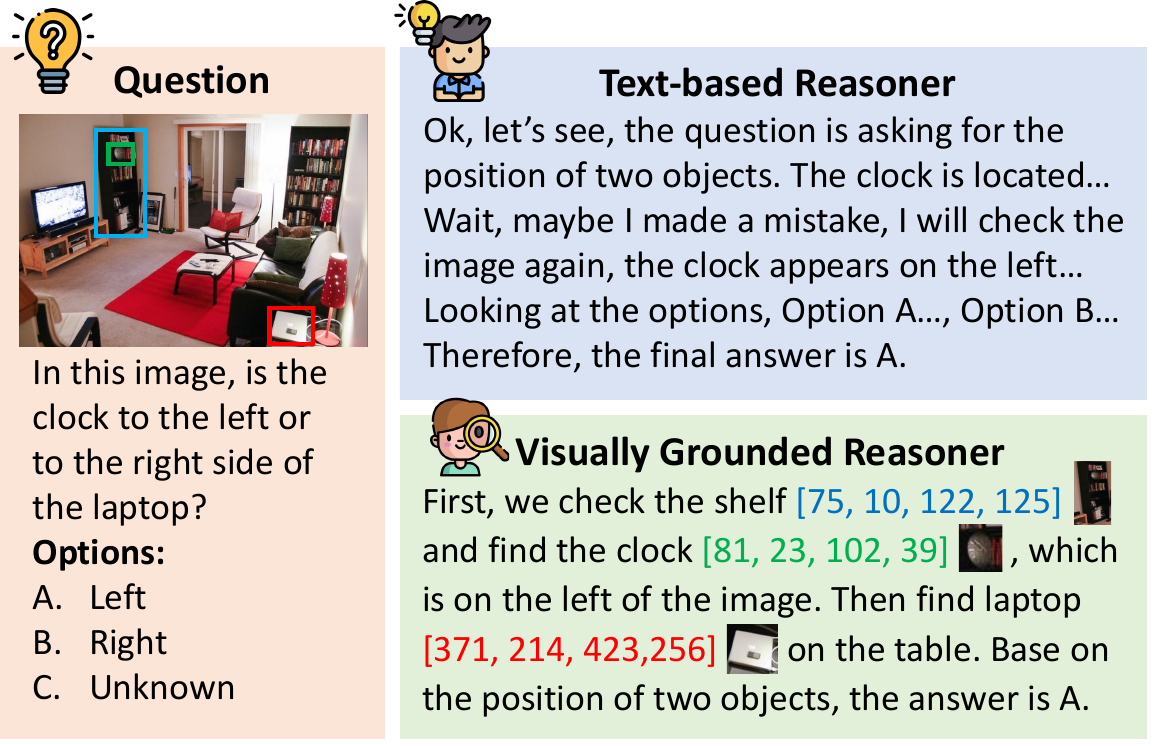} 
    \caption{
    Rollout examples of two reasoning modes.}
    \label{fig:intro-a}
  \end{subfigure}
  \hfill
  \begin{subfigure}{0.51\linewidth}
  \centering
    \includegraphics[width=\linewidth]{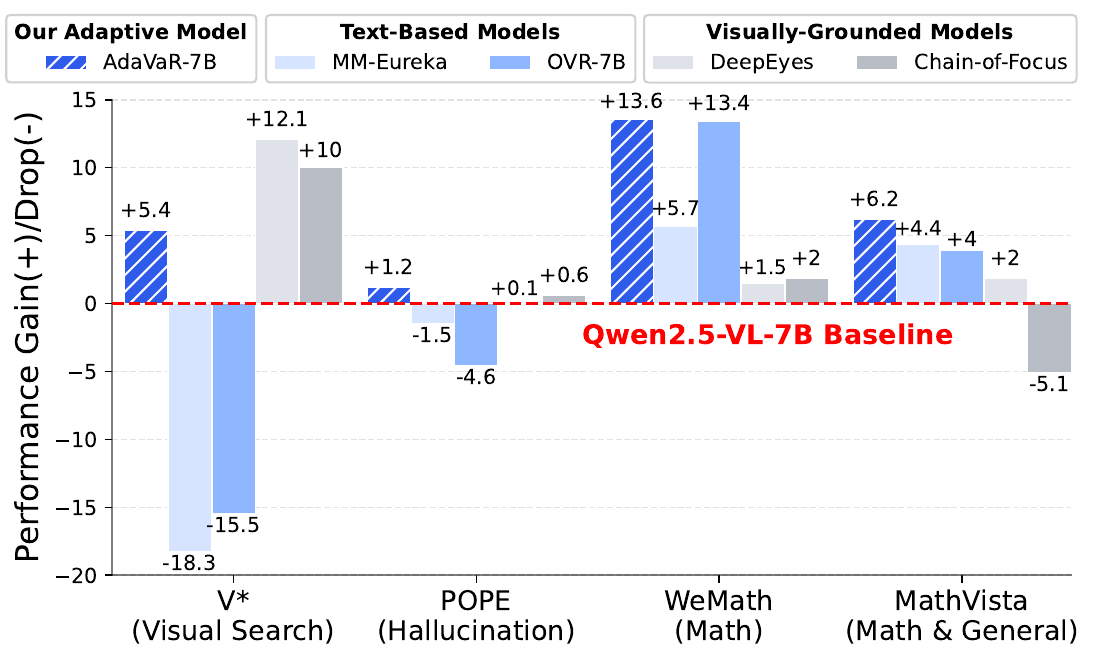}
    \caption{Performance of different reasoning modes.}
    \label{fig:intro-b}
  \end{subfigure}
  \vspace{-2mm}
  \caption{Comparison between two reasoning modes. (b) Performance gains (positive values) and drops (negative values) of Qwen2.5-VL-7B–based reasoning models relative to the base model.}
  %, including the text-based MM-Eureka and OVR-7B, the visually grounded DeepEyes and Chain-of-Focus, and our proposed AdaVaR-7B.}
  \label{fig:main-figure}
  \vspace{-4mm}
\end{figure}

% Motivated by these advancements, considerable effort has been paid to extending the success of reasoning techniques to the multi-modal domain.
% aiming to enhance the comprehension and reasoning abilities of large vision-language models (LVLMs) in complex visual-textual contexts.
Motivated by these advancements, preliminary explorations of reasoning in multi-modal contexts have emerged.
As illustrated in Figure~\ref{fig:intro-a}, existing visual reasoning methods fall into two types based on the form of CoTs (referred to as ``reasoning mode'' in this paper): (1) Text-based reasoning, which is consistent with LLMs, where large vision-language models (LVLMs) represent the reasoning process directly in natural language~\citep{yang2025r1onevisionadvancinggeneralizedmultimodal,huang2025visionr1incentivizingreasoningcapability,meng2025mmeurekaexploringfrontiersmultimodal}; (2) Visually-grounded reasoning, which anchors the reasoning process in the visual context. Typically, this involves guiding LVLMs to generate structured outputs (e.g., bounding boxes or points) that align textual concepts in the reasoning process with specific image regions~\citep{chen2023shikra,lei2024scaffolding,fan2025gritteachingmllmsthink}. Additionally, several works propose to introduce associated local visual information into the generated sequence, enabling the model to focus on specific regions for subsequent inference~\citep{li2025vocotunleashingvisuallygrounded,shao2024visual,zheng2025deepeyesincentivizingthinkingimages,o3}.

% Different reasoning modes also lead to differences in performance. 
% Different reasoning modes of LVLMs exhibit distinct strengths and inherent trade-offs rooted in their core design paradigms.
% The inductive biases inherent in different reasoning modes lead to varying strengths and weaknesses across domains.
% As illustrated in Figure~\ref{fig:intro-b}, text-based models demonstrate superiority in abstract reasoning tasks (e.g., geometric problem-solving), but are more prone to hallucination issues. In contrast, visually grounded models achieve reliable hallucination control and excel in scenarios where object information is clearly defined, showing limited performance gains on mathematical reasoning benchmarks. 
% This indicates that no single reasoning mode currently dominates across all task scenarios, which naturally raises a critical question: how can we integrate different reasoning modes into a unified framework, thereby enabling the model to adaptively leverage their advantages across diverse scenarios?
% These phenomena indicate that models limited to a specific reasoning mode tend to become experts in particular domains, while struggling to develop general visual reasoning abilities.

% 不同的模式各有优势
% The inductive biases inherent in different reasoning modes lead to varying strengths and weaknesses across domains. 
Different reasoning modes impose distinct inductive biases, yielding varying strengths and weaknesses across domains.
As shown in Figure~\ref{fig:intro-b}, text-based models excel at abstract reasoning (e.g., mathematics) but are more prone to hallucinations due to overthinking and language bias. In contrast, visually-grounded models are better at leveraging visual information, curbing hallucination, and handling problems with clear object information, yet show limited gains on mathematical benchmarks (since abstract concepts like length, size, etc. cannot be grounded to provide useful information). 
Overall, no single mode dominates across all tasks. This raises a critical question: can we integrate the complementary strengths of different modes to build a general visual reasoning model?

% As presented in Figure~\ref{fig:intro-b}, each reasoning approach excels in different domains: text-based reasoning methods are well-suited for abstract mathematical problems, while grounded reasoning is more appropriate for scenarios where object information is clearly defined.

% Building on these insights, we argue that an ideal visual reasoning models should not only enhance its reasoning abilities within a specific paradigm, but also adaptively select the appropriate reasoning mode based on the question. However, learning such adaptive reasoning capabilities presents several challenges: (1) it is difficult to incentivize the adaptive selection mechanism in a supervised manner, as the corresponding supervision signals are hard to annotate; (2) different reasoning modes typically require specific input prompts for guidance, resulting in input discrepancies across modes; and (3) existing visual reasoning models mainly undergo a supervised fine-tuning (SFT) phase for cold start, which tends to bias the model toward a particular reasoning paradigm and weakens its exploration of different modes during the reinforcement learning process.

% In light of this, we propose a novel paradigm for visual reasoning: many minds in one brain. Similar to how humans tackle various problems with different ideas, we argue that intelligent visual reasoning models should be able to unify different reasoning modes and adaptively select the appropriate mode based on the context, thereby exhibiting general-purpose reasoning capabilities.
In light of this, we propose a novel visual reasoning paradigm, Mixture-of-Visual-Thoughts: the model is able to reason in different modes and adaptively choose the appropriate one based on the context.
However, the greatest challenges lie in 
% (i) how to uniformly represent and enable the model to learn different reasoning modes within a unified model, 
(i) how to uniformly represent different reasoning modes and enable a unified model to learn them;
and (ii) how to develop the capability for context-adaptive mode selection.
To tackle these challenges, we introduce \textbf{AdaVaR}, an \underline{Ada}ptive \underline{V}isu\underline{a}l \underline{R}easoning framework with a two-stage training paradigm: a supervised cold-start learning of different modes, and an adaptive RL to allow LVLMs to explore and acquire mode selection skills.

Specifically, AdaVaR begins with a cold-start stage to unify multiple reasoning modes within a model. 
We define a uniform sequence format for the reasoning paths of different modes and distinguish them with mode-specific prefix tokens. This allows us to mix reasoning data across modes and train the model through a unified supervised fine-tuning (SFT) stage.
% We unify the reasoning paths of different modes into a single sequence, distinguished by introduced mode-specific prefix tokens. Then we directly mix reasoning data across modes and train the model through a unified supervised fine-tuning (SFT) stage. 
% We uniformly define reasoning paths of different modes as a sequence consisting of: a mode prefix token, reasoning process, and answer, with each mode assigned a specific prefix token to distinguish them within a unified representation. 
% Leveraging the autoregressive property, this sequence order naturally enables the model to first select a mode (by generating prefix tokens) and then proceed with reasoning, all within a single generation process. Facilitated by the unification, reasoning data of different modes can be directly mixed up, allowing the model to learn and differentiate various reasoning modes through an unified supervised fine-tuning (SFT) stage.  
% However, since we cannot pre-estimate the relative strengths between different modes for various questions in advance, it is difficult to construct supervision signals during the SFT stage to guide appropriate mode selection.
% Apart from learning different reasoning modes,
Although different reasoning modes can be effectively learned,
guiding the selection of appropriate modes via SFT remains challenging, since we cannot pre-estimate the relative performance of different modes for each problem.
% it is difficult to guide appropriate mode selection during the SFT stage, since we cannot estimate in advance the relative performance of different modes on each problem.

% For the second challenge, 
% we use RL to strengthen reasoning and induce adaptive mode selection. 
Therefore, to address the second challenge, a subsequent RL stage is introduced to induce context-adaptive mode selection skills while enhancing the reasoning abilities.
%The idea is intuitive: the policy model is guided to explore different modes for the same question, and receives reinforced supervision to modes that are more likely to yield correct answers.
The idea is straightforward: we encourage the model to explore different modes for each question and reward the mode that is more likely to yield correct answers.
Nonetheless, standard GRPO~\citep{shao2024deepseekmath} poses obstacles to achieving this goal. 
% First, the policy model may suffer from insufficient mode exploration--for instance, it only generates rollouts from a single mode when tackling one question. 
First, the policy model may under-explore reasoning modes without explicit guidance (e.g., producing rollouts from only one mode per question).
Second, the rollout-level advantage in GRPO fails to explicitly capture preferences among different modes.
To address these issues, we propose AdaGRPO, which steers exploration and comparison between reasoning modes through three key components: 
(1) Prefix-guided mode exploration: by fixing the mode prefix, we compel the policy model to explore evenly across multiple modes for the same sample. 
% (2) We design an advantage calculation strategy to explicitly guide mode selection and enhance reasoning ability; 
(2) Adaptive advantage mechanism: besides the rollout-level advantage, we introduce a mode-relative advantage to explicitly guide the model towards optimal mode selection.
(3) Curriculum-based data scheduling: we initiate training with easier data to learn coarse-grained distinctions between modes, then gradually transition to harder questions for fine-grained mode selection capability.

%Based on the AdaVaR framework, we conduct extensive experiments on top of Qwen2.5-VL~\citep{bai2025qwen25vltechnicalreport}. As shown in Figure~\ref{fig:intro-b}, the experimental results demonstrate that our method effectively enable the learning of different reasoning modes and context-adaptive mode selection capabilities, ultimately exhibiting general visual reasoning abilities: achieving consistent improvements across various scenarios. In summary, the main contribution of this paper is three-fold:
Extensive experiments demonstrate that AdaVaR enables the learning of different reasoning modes and context-adaptive mode selection capabilities. As shown in Figure~\ref{fig:intro-b}, our AdaVaR-7B achieves consistent improvements across scenarios. In terms of average performance over 8 various benchmarks, AdaVaR-3B matches Qwen2.5-VL-7B, and AdaVaR-7B surpasses GPT-4o. Our code, models, and data will be open-sourced to the community. The contributions of this work are threefold:
\begin{itemize}
    \item Unlike existing works that focus on specific reasoning modes, we propose a novel visual reasoning paradigm, MoVT: unifying different modes within a model and guiding it to select the appropriate one based on context, pushing the frontier of general visual reasoning.
    \item We introduce AdaVaR, a two-stage adaptive reasoning training framework: different modes are unified and learned during the supervised cold-start stage, and adaptive selection of appropriate modes is learned during the RL process using the proposed AdaGRPO algorithm.
    \item Extensive experiments demonstrate the effectiveness of AdaVaR in guiding LVLMs to learn and distinguish different reasoning modes in a unified manner, and to dynamically select modes based on context. Our adaptive reasoning method brings general improvements across various scenarios and exhibits considerable potential for future exploration.
\end{itemize}

\section{Related Works}

% \subsection{Large Vision-Language Models}

% \paragraph{Construction of LVLMs} Inspired by the success of LLMs, researchers have explored to construct LVLMs by encoding images and videos with visual encoders~\citep{radford2021learning,zhai2023sigmoid,dehghani2023patch} and mapping them into the input space of LLMs through different types of connection modules~\citep{alayrac2022flamingo,li2023blip,liu2023improved}. After subsequent pre-training, SFT, and optional RL phases, LVLMs are able to understand multi-modal context and respond adhere to the instructions~\citep{liu2024llava,li2024llavaonevisioneasyvisualtask,bai2025qwen25vltechnicalreport,zhu2025internvl3}.
% \paragraph{Grounded LVLMs} To capture fine-grained local details, recent work strengthens the grounding abilities of LVLMs. The prevailing approach is to represent bounding boxes using text~\citep{bai2023qwen,bai2025qwen25vltechnicalreport,chen2023minigpt} or special tokens~\citep{peng2023kosmos}, which are embedded into the input and output sequences, training LVLMs to align coordinates with corresponding regions. In addition, extra input feature extractor~\citep{you2023ferret,zhang2024gpt4roi,yuan2024osprey} and output modules~\citep{zhang2024llava,lai2024lisa,rasheed2024glamm} can be introduced to enable LVLMs to understand and generate pixel-level information, such as segmentation maps.

% \subsection{Reasoning Models}

\subsection{Language Reasoning Models}
% 文本的推理模型，从早期的 CoT 到后来如何用强化学习提升推理能力。
Early work reveals the reasoning potential of LLMs, well-designed prompts~\citep{kojima2022large} and in-context samples~\citep{wei2022chain} can be adopted to guide LLMs to generate CoTs, enhancing the ability to solve complex questions. Subsequent research focuses on designing complex forms of CoT~\citep{yao2024tree,besta2024graph} and introducing mechanisms like majority voting~\citep{wang2023selfconsistency}, reflection~\citep{madaan2023self,shinn2023reflexion}, and step-level search~\citep{lample2022hypertree,xie2023self,tian2024toward}. Recently, works represented by DeepSeek-R1~\citep{guo2025deepseek} have demonstrated that the reasoning abilities of LLMs can be effectively incentivized and expanded under a highly scalable RL environment~\citep{qwq32b,yu2025dapoopensourcellmreinforcement}.

% 目前文本的自适应推理主要应用在切换长/短/不思考的角度
% \paragraph{Adaptive Reasoning in LLMs} To balance performance and efficiency, recent work explores adaptive reasoning methods that switch between thinking and non-thinking based on the question difficulty, using techniques like prompting~\citep{ma2025reasoning,muennighoff2025s1} and RL~\citep{zhang2025adaptthink,yang2025think}. 
% In contrast, we focus on the ability to adaptively select between different reasoning modes with complementary strengths, achieving general-purpose improvement.
% Models are guided to decide whether to reason based on the difficulty of the question, as well as to streamline the reasoning processes.

\subsection{Vision-Language Reasoning Models}

\paragraph{Text-Based Reasoning Models} Mainstream visual reasoning models directly express thinking processes in natural language. Early methods first learn reasoning patterns via SFT, building datasets through complex distillation~\citep{xu2024llava,thawakar2025llamav} or collective search frameworks~\citep{yao2024mulberry}, and then enhance the reasoning capability with search-based mechanisms~\citep{xu2024llava,thawakar2025llamav} or DPO~\citep{zhang2024improve,dong2025insight}. Recent works mainly follow DeepSeek-R1, using high-quality data to compute verifiable rewards and employing GRPO to incentivize the reasoning abilities~\citep{liu2025visual,ma2025one,peng2025lmm,meng2025mmeurekaexploringfrontiersmultimodal,xu2025geosense}. Meanwhile, A cold-start SFT phase before RL, trained on high-quality data, can bootstrap the model’s reasoning and enable more effective exploration during RL~\citep{yang2025r1onevisionadvancinggeneralizedmultimodal,huang2025visionr1incentivizingreasoningcapability,wei2025open,chen2025sft}.

\paragraph{Visually-Grounded Reasoning Models} Another line of research proposes to explicitly ground the reasoning process onto images. The prevailing approach is to introduce bounding box coordinates in CoTs, either through prompting~\citep{mitra2024compositional} or SFT~\citep{chen2023shikra,wu2025grounded}. Furthermore, the located regions can be cropped, zoomed, and then fed back into the model to assist it in referring to fine-grained visual information for subsequent reasoning~\citep{li2025vocotunleashingvisuallygrounded,shao2024visual,shen2024zoomeye,o3}. In addition, tool-based~\citep{zhou2024image} and visual prompting~\citep{lei2024scaffolding,yang2023setofmark} methods can be used to supplement images with coordinate systems or segmentation maps, which can be utilized by LVLMs for explicit grounding.
Recent works further demonstrate that grounded reasoning abilities can be activated and enhanced through RL by designing rewards~\citep{sarch2025grounded,fan2025gritteachingmllmsthink} and methods to guide grounded rollout generation~\citep{cao2025ground,zheng2025deepeyesincentivizingthinkingimages}.

% 本文的特点
Existing models typically focus on one of the above modes and a narrow domain: text-based reasoning targets mathematical problems, while the visually-grounded mode is mainly used for object-centric tasks. Inspired by this, we propose an adaptive reasoning paradigm, MoVT, that integrates the complementary strengths of different modes to build a general-purpose visual reasoning model.
% Inspired by the strengths of the two reasoning modes in their respective domains, this paper aims to integrate different modes and leverage their complementary strengths to build a general visual reasoning model.

\section{AdaVaR: An Adaptive Visual Reasoning Framework}
\label{section:method}

This section details how the proposed AdaVaR framework endows LVLMs with adaptive reasoning abilities. First, to integrate multiple reasoning modes in a unified model, we introduce a reasoning paradigm in \S~\ref{section:mode_unification} to represent reasoning paths from different modes in a unified manner, merging mode selection and reasoning into a single autoregressive generation process. Building on this, the model undergoes an SFT phase to learn multiple reasoning modes simultaneously (as described in \S~\ref{section:SFT_stage}). Based on the SFT model, we design an RL algorithm (as detailed in \S~\ref{section:RL_stage}) to help the model acquire the ability to select the appropriate mode according to the provided context.

\subsection{Reasoning Mode Definition and Unification}
\label{section:mode_unification}

As introduced in the introduction, we regard different forms of CoT as distinct reasoning modes. In this paper, we mainly consider the integration of two commonly applied visual reasoning modes: (i) \textbf{Text-based reasoning}, without introducing additional inductive bias, represents all reasoning processes only in language; (ii) \textbf{Visually-grounded reasoning} requires models to generate structural information, such as coordinates, to align key concepts--primarily objects--with corresponding regions in the image during reasoning. In this work, we adopt the widely used bounding-box-enhanced format~\citep{li2025vocotunleashingvisuallygrounded,fan2025gritteachingmllmsthink}, guiding the model to describe objects in the form of ``object $[x_1, y_1, x_2, y_2]$''\footnote{Absolute coordinates of the top-left and bottom-right corners, which are aligned with Qwen2.5-VL.} to explicitly align the reasoning process with the image.

% Please note that in this paper we mainly explore the unification and adaptive selection across two common modes: text-based reasoning and grounded reasoning. However, the AdaVaR framework can be naturally extended to more modes.

\paragraph{Reasoning Mode Unification} To enable the model to learn and distinguish different modes within a unified framework, we define a uniform format of reasoning sequences: each mode is assigned a unique prefix token, which is placed at the beginning of reasoning paths to serve as an in-context indicator, guiding the model to differentiate between modes. We define the reasoning template as:
\begin{center}
% \vspace{-0.2cm}
    \fcolorbox{black}{gray!10}{\parbox{.95\linewidth}{\textbf{System}: You are a helpful assistant. You need to first think about the reasoning process and then answer the question raised by the user, in the format of \textless think\textgreater\;reasoning process here \textless/think\textgreater\; \textless answer\textgreater\;answer here \textless/answer\textgreater. You have two modes of thinking: \\
    1. \textit{Grounded Reasoning}: ..., begin your response with \textcolor{blue}{\textless ground\textgreater}\;when using this mode.\\
    2. \textit{Text-based Reasoning}: ..., begin your response with \textcolor{blue}{\textless text\textgreater}\;when using this mode. \\
    \textbf{User}: prompt. \textbf{Assistant}: \textcolor{blue}{\textless mode prefix\textgreater} \textcolor{red}{\textless think\textgreater\;... \textless/think\textgreater\; \textless answer\textgreater\;... \textless/answer\textgreater}}}
    % \vspace{-0.2cm}
\end{center}
The system prompt specifies the reasoning requirements and introduces the different reasoning modes (omitted for brevity; the full prompt is provided in Appendix~\ref{appendix:detailed_prompts}). The model response includes the \textcolor{blue}{mode prefix} and the \textcolor{red}{reasoning process}. Leveraging the autoregressive property, generating reasoning paths in this format can be naturally divided into two steps, i.e., $P(a,t,m|i,q)=P(m|i,q)\times P(a,t|m,i,q)$: (1) $P(m|i,q)$ selecting the reasoning mode by generating the mode prefix $m$ based on the input image $i$ and question $q$, and (2) $P(a,t|m,i,q)$ generating the corresponding thinking process $t$ and answer $a$ based on the selected mode $m$ and the input context $i,q$. These two steps can be accomplished sequentially within a single sequence generation process.

\subsection{Stage 1: Cold-Start Mode Learning with SFT}
\label{section:SFT_stage}
Building on the unified reasoning format, we adopt an SFT stage as a cold start to enable the base model, Qwen2.5-VL~\citep{bai2025qwen25vltechnicalreport}, to learn reasoning abilities across different modes simultaneously. To this end, we mix expert reasoning trajectories from different modes: (i) For text-based reasoning, we follow DeepSeek-R1~\citep{guo2025deepseek} to construct data by distilling a text-based visual reasoning model~\citep{ma2025one} and applying rejection sampling. (ii) Regarding visually-grounded reasoning, we directly utilize high-quality SFT data constructed in existing works~\citep{li2025vocotunleashingvisuallygrounded}. The proportion of the two parts of data is controlled to 1:1 to prevent introducing undesired bias on mode selection. Please refer to Appendix~\ref{appendix:train_data} for more details.

\subsection{Stage 2: Adaptive Mode Selection with RL}
\label{section:RL_stage}

\begin{figure}[t]
    \centering
    \includegraphics[width=0.98\linewidth]{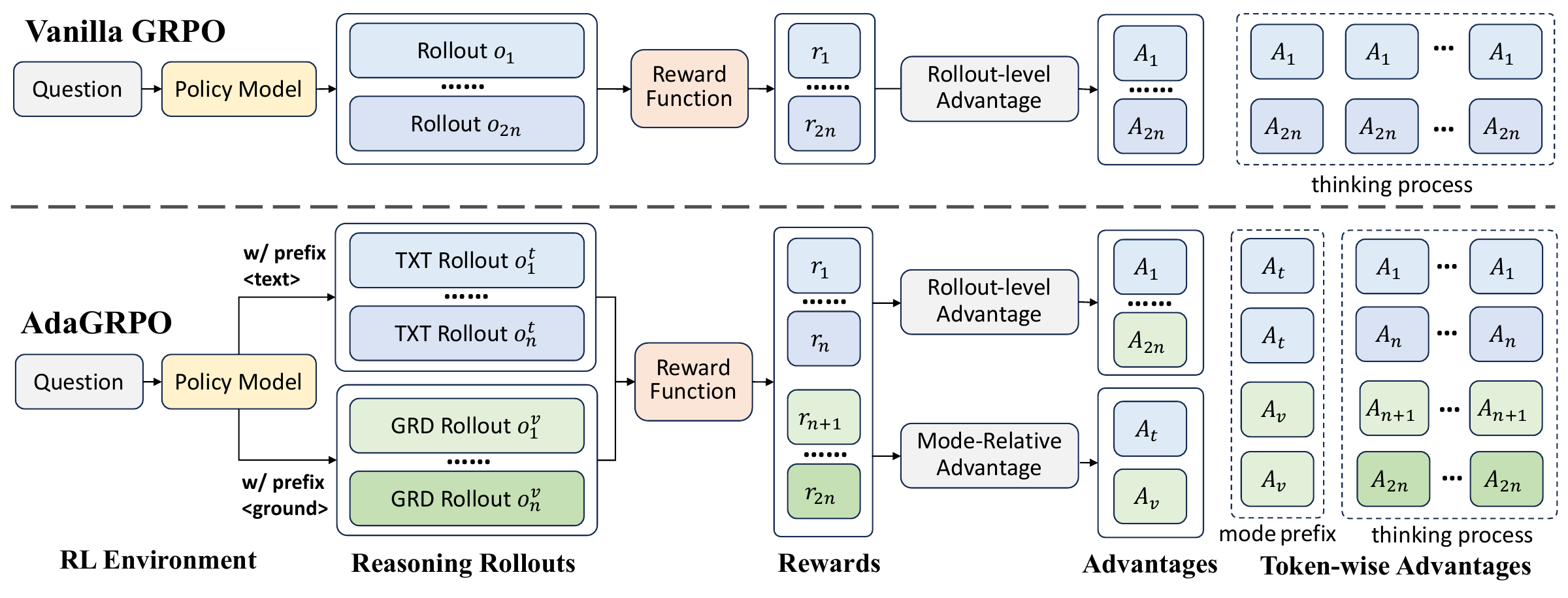}
    \vspace{-2mm}
    \caption{Demonstration of GRPO and AdaGRPO. In AdaGRPO, we use mode prefixes to guide exploration across different modes (TXT and GRD respectively represent text-based and visually-grounded reasoning), calculate both rollout-level advantages $\{A_{j}\}_{j=1}^{2n}$ and mode-relative preference $\{A_t,A_v\}$, and explicitly guide mode selection and the enhancement of reasoning abilities.}
    \label{fig:AdaGRPO}
\end{figure}
\vspace{-2mm}

Although Stage 1 helps the model learn reasoning abilities in different modes, it is difficult to learn mode selection under the supervised learning paradigm because we cannot pre-construct effective supervision on which mode to select. Moreover, recent studies have shown that the reasoning abilities acquired during the SFT stage can be further enhanced through RL~\citep{yang2025r1onevisionadvancinggeneralizedmultimodal,wei2025open}. Therefore, we introduce an RL stage, which simultaneously improves the reasoning ability and guides the model to learn adaptive mode selection based on the proposed AdaGRPO algorithm.

\subsubsection{RL Algorithm: AdaGRPO}
\label{section:adagrpo}
\paragraph{Revisiting GRPO} For a better illustration on the design rationale of AdaGRPO, we review the details of GRPO~\citep{shao2024deepseekmath}. As shown in the upper part of Figure~\ref{fig:AdaGRPO}, the old policy model $\pi_{\theta_{old}}$ generates $2n$ reasoning rollouts $\{o_1,o_2,...,o_{2n}\}$ for question $q$, obtaining rewards $\{r_1,r_2,...,r_{2n}\}$ which are utilized to optimize the current policy model $\pi_{\theta}$ with the following objective: 
\begin{align}
\label{equation:vanilla_grpo}
    &\mathcal{J}_{GRPO}(\theta)=\mathbb{E}\left[q\sim P(Q),\{o_j\}_{j=1}^{2n}\sim\pi_{\theta_{old}}(O|q)\right]\nonumber\\ 
    &\frac{1}{2n}\sum_{j=1}^{2n}\frac{1}{|o_j|}\sum_{t=1}^{|o_j|}\left(\min\left[\frac{\pi_{\theta}(o_j|q,o_{j,<t})}{\pi_{\theta_{old}}(o_j|q,o_{j,<t})}A_{j,t},\text{clip}\left(\frac{\pi_{\theta}(o_j|q,o_{j,<t})}{\pi_{\theta_{old}}(o_j|q,o_{j,<t})}, 1-\epsilon,1+\epsilon\right)A_{j,t}\right]\right)
\end{align}
where $\epsilon$ is the clipping hyper-parameter and the KL penalty is omitted for brevity. $A_{j,t}=A_j=\frac{r_j-\text{mean}(\{r_1,...,r_{2n}\})}{\text{std}(\{r_1,...,r_{2n}\})}$ are the rollout-level advantages estimated by comparing rewards of all rollouts. 

Although effective, the vanilla GRPO is limited in the adaptive reasoning scenarios in terms of: (i) the policy model, especially after SFT, may exhibit preferences for certain reasoning modes. As a result, the generated $2n$ rollouts could come from the same mode, leading to uneven exploration on different reasoning modes; (ii) GRPO merely considers rollout-wise advantages without explicit modeling on preference between different reasoning modes to guide mode selection. 

% \subsubsection{RL Algorithm: AdaGRPO}

Building on the insights above, we propose AdaGRPO, a variant of GRPO that encourages exploration across reasoning modes and guides mode selection with mode-relative advantage estimation.

\paragraph{Prefix-Guided Mode Exploration} First, AdaGRPO enforces uniform exploration on different modes during the rollout sampling process. The unified reasoning format defined in Section~\ref{section:mode_unification} allow us to guide the model to perform reasoning in a specific mode by fixing the corresponding mode prefix in the input sequence. We divide the $2n$ rollouts into two sub-groups: $\{o_j^t\}_{j=1}^n\sim\pi_{\theta_{old}}(O|i,q,m_t)$ consists of $n$ text-based reasoning rollouts with prefix $m_t=\textless\text{text}\textgreater$, and $\{o_j^v\}_{j=1}^n\sim\pi_{\theta_{old}}(O|i,q,m_v)$ consists of $n$ visually-grounded rollouts with prefix $m_v=\textless\text{ground}\textgreater$.

\paragraph{Reward Function} We follow the reward setup utilized in DeepSeek-R1~\citep{guo2025deepseek}, which includes format rewards and accuracy rewards. The former controls the output format, while the latter assigns a reward of 1 or 0 based on rule-based evaluation of the produced answer. The same reward functions are applied to all rollouts from different modes $\{o_1^t,...,o_n^t,o_1^v,...,o_n^v\}$ to support subsequent computation of advantages both between reasoning modes and among rollouts.

\paragraph{Adaptive Advantage Calculation} To explicitly guide the mode selection, we characterize the mode-relative advantage based on the rewards $\{r_j\}_{j=1}^n$ and $\{r_j\}_{j=n+1}^{2n}$, corresponding to the two sub-groups of rollouts $\{o_j^t\}_{j=1}^n$ and $\{o_j^v\}_{j=1}^n$. Following the Gaussian assumption in GRPO, we obtain two normal distributions of rewards for the two sub-groups, $\{r_j\}_{j=1}^n\sim P_t$ and $\{r_j\}_{j=n+1}^{2n}\sim P_v$, then we estimate the advantage of mode A over mode B with the probability that a rollout sampled from mode A outperforms a rollout sample from mode B (and vice versa), formulated as:
\begin{align}
    A_v=&P_{X\sim P_v,Y\sim P_t}(X>Y);\;\;\; A_t=P_{X\sim P_v,Y\sim P_t}(X<Y) \nonumber\\
    \text{where}\;\;P_t&=N\left(\mu_t=\text{mean}(\{r_j\}_{j=1}^n),\sigma_t^2=\text{Var}(\{r_j\}_{j=1}^n)\right)\nonumber\\
    P_v&=N\left(\mu_v=\text{mean}(\{r_j\}_{j=n+1}^{2n}),\sigma_v^2=\text{Var}(\{r_j\}_{j=n+1}^{2n})\right)
\end{align}
Furthermore, we can reasnonably assume that $X$ and $Y$ are independent, so $X-Y\sim N(\mu_v-\mu_t,\sigma_v^2+\sigma_t^2)$. In this way, $A_v$ and $A_t$
 can be directly calculated with the cumulative distribution function (CDF) $\Phi$ of the standard Gaussian distribution as $A_v=\Phi(\frac{\mu_v-\mu_t}{\sqrt{\sigma_v^2+\sigma_t^2}})=1-A_t$.
 
 As in GRPO, we also compute rollout-wise advantages $\{A_j\}_{j=1}^{2n}$. However, unlike GRPO, we assign different advantages to different tokens within the same sequence. As illustrated in the lower part of Figure~\ref{fig:AdaGRPO}, mode-relative advantages, $A_t$ and $A_v$, are assigned to mode prefix tokens, guiding models to select the better mode. Rollout-wise advantages $\{A_j\}_{j=1}^{2n}$ are applied to tokens in the thinking processes, enhancing the reasoning capability of the policy model. The specific formulation is:
 \begin{align}
 A_{j,t}'=
 \begin{cases}
 \mathbb{1}_{\{o_j\in\{o_k^v\}_{k=1}^n\}}A_v+\mathbb{1}_{\{o_j\in\{o_k^t\}_{k=1}^{n}\}}A_t\;\;\;\; &\text{if  }o_{j,t}\in m \\
 A_j=\frac{r_j-\text{mean}(\{r_1,r_2,...,r_{2n}\})}{\text{std}(\{r_1,r_2,...,r_{2n}\})} &\text{otherwise}
 \end{cases}
 \end{align}

In summary, the optimization objective of the proposed AdaGRPO is formulated as:
\begin{align}
    &\mathcal{J}_{AdaGRPO}(\theta)=\mathbb{E}\left[i,q\sim P(I,Q),\{o_j\}_{j=1}^{n}\sim\pi_{\theta_{old}}(O|q,i,m_t),\{o_j\}_{j=n+1}^{2n}\sim\pi_{\theta_{old}}(O|q,i,m_v)\right]\nonumber\\ 
    &\frac{1}{2n}\sum_{j=1}^{2n}\frac{1}{|o_j|}\sum_{t=1}^{|o_j|}\left(\min\left[\frac{\pi_{\theta}(o_j|i,q,o_{j,<t})}{\pi_{\theta_{old}}(o_j|i,q,o_{j,<t})}A'_{j,t},\text{clip}\left(\frac{\pi_{\theta}(o_j|i,q,o_{j,<t})}{\pi_{\theta_{old}}(o_j|i,q,o_{j,<t})}, 1-\epsilon,1+\epsilon\right)A'_{j,t}\right]\right) \nonumber
\end{align}

\subsubsection{Dataset Collection}
\label{section:rl_data}
For RL, we collect a diverse dataset consisting of two parts: (1) existing datasets with verifiable answers, including Geo170K~\citep{gao2023gllava}, OmniCount~\citep{mondal2025omnicount}, and MM-Eureka~\citep{meng2025mmeurekaexploringfrontiersmultimodal}; (2) a curated subset from the SFT data of LLaVA-OneVision~\citep{li2024llavaonevisioneasyvisualtask} and InternVL~\citep{chen2024far}, focusing on math, OCR, object counting, science, grounding, and document-related tasks. The subset is filtered based on the answer verifiability and difficulty, and down-sampled to balance different tasks. Please refer to Appendix~\ref{appendix:train_data} for more details.
% that covers various scenarios, features answers that can be judged by rules, and has been filtered according to difficulty.

\paragraph{Curriculum Learning} 
To help the model progressively learn reasoning and mode selection in complex scenarios, we design two data mixing strategies: binary mixture, which includes only the OmniCount and Geo170K (with Geo170K being relatively simple geometry problems); and diverse mixture, which consists of all remaining data beyond binary mixing, covering multiple tasks and featuring higher difficulty. Our model first learns from the binary mixed data, then from the diverse mixed data, with both the difficulty and task distribution progressing from simple to complex.
% In addition, since AdaGRPO entangles the learning of mode selection with reasoning enhancement, it is challenging for models in complex scenarios. To address this, we propose a curriculum learning-based data sampling strategy. Specifically, instead of sampling question uniformly as in Equation~\ref{equation:vanilla_grpo}, we sample according to a dynamic distribution $P_c$: In the early stage of training, we sample questions only from GeoQA and OmniCount, guiding the model to solve simple tasks and learn coarse-grained mode selection (distinguishing between distinctly different scenarios). In the later stage, questions are sampled from all datasets, allowing the model to learn generalized reasoning capabilities and finer-grained mode selection between more similar scenarios.

\section{Experiments}

\subsection{Experimental Setup}
\label{section:experimental_setup}

\paragraph{Implementation Details} Based on the proposed framework, we develop two reasoning models, AdaVaR-3B and AdaVaR-7B, which are constructed from Qwen2.5-VL-3B and Qwen2.5-VL-7B~\citep{bai2025qwen25vltechnicalreport}, respectively. Please refer to Appendix~\ref{appendix:training_details} for detailed hyperparameter settings.

\paragraph{Evaluation Benchmarks} Unlike existing works that focus on specific domains, we comprehensively evaluate the reasoning abilities of AdaVaR across diverse scenarios. For mathematical reasoning, we consider MathVista~\citep{lu2023mathvista}, MathVision~\citep{wang2024measuringmultimodalmathematicalreasoning}, MathVerse~\citep{zhang2024mathversedoesmultimodalllm}, and WeMath~\citep{qiao2024we}. In additional, for general-purpose scenarios, we adopt V*~\citep{wu2023textit}, SpatialScore~\citep{wu2025spatialscoreunifiedevaluationmultimodal}, and MMStar~\citep{chen2024we} to evaluate the capabilities of models in visual search, spatial reasoning, and general perceptual reasoning, respectively, while POPE~\citep{li2023evaluating} is used to diagnose hallucination risk.

During evaluation, our AdaVaR model naturally incorporates a mode-switching mechanism: if the model gets stuck in repetitive logic and cannot conclude an answer (between \textless answer\textgreater~and \textless /answer\textgreater) while reasoning in a particular mode, we switch to another mode and try again. More evaluation details, including descriptions of the compared models, are included in Appendix~\ref{appendix:evaluation_details}.

\input{tables/main_table}

\subsection{Overall Performance}

Table~\ref{table:main_results} presents a thorough evaluation of visual reasoning models. Several insights can be gleaned: (1) As argued in Section~\ref{section:intro}, models based on a single reasoning mode tend to become experts in specific domains. Text-based models, such as VLAA-Thinker-3B and OVR-7B, achieve notable improvement on mathematical problems but show significant degradation in V* and POPE. In contrast, visually-grounded models consistently improve performance on V* and control hallucination, but struggle with math problems--DeepEyes makes a slight improvement while other methods fail to maintain the original mathematical capability of Qwen2.5-VL. (2) AdaVaR-3B and AdaVaR-7B, on the other hand, combine the strengths of two modes in an adaptive manner, and are the only models that outperform Qwen2.5-VL across all datasets. Specifically, AdaVaR achieves the best performance on MathVista, WeMath, and POPE, while excelling on MMStar, MathVision, and V*. (3) In terms of the overall performance measured by average accuracy, AdaVaR achieves the best results in both the 3B and 7B groups. Notably, AdaVaR-3B reaches a level very close to Qwen2.5-VL-7B, while AdaVaR-7B surpasses GPT-4o.
%, second only to OVR-7B, which holds an overwhelming advantage in mathematics. 
In summary, our adaptive reasoning framework provides an effective and feasible solution for building general visual reasoning models.

\subsection{Delving into Adaptive Visual Reasoning}
\label{section:delving}

Beyond the overall improvement, we delve into the mechanism of adaptive reasoning based on AdaVaR-3B and several ablated models, aiming to answer the questions raised in Section~\ref{section:intro}.

\input{tables/delve_adaptive_v2}

\subsubsection{Question 1: Can we unify different reasoning modes in one model?} 
\label{section:delving_question_1}

As shown in Table~\ref{table:adaptive_analysis}, after SFT and RL, two reasoning modes induce different prediction patterns in terms of performance: the text-based mode excels at math, while the grounded mode is better at object-centric tasks. 
% Cases in Appendix~\ref{appendix:case_study} also demonstrate the different characteristics of two modes.
% Please see cases from different modes in Appendix~\ref{appendix:case_study}.
% Appendix~\ref{appendix:case_study} provides qualitative comparison between two modes.
Please also see cases in Appendix~\ref{appendix:case_study} for a qualitative comparison.
% 接一下 case study

\paragraph{Comparison with Single-Mode Baselines} Furthermore, we construct two single-mode baselines, Grounded-SFT-RL and Text-SFT-RL, under the same training settings: using SFT data corresponding to each mode and RL data consistent with AdaVaR. We find that: (1) The difference between the single-mode baselines and the respective modes of AdaVaR is minimal, suggesting that consolidating both modes into one model does not hinder the improvement of either individual mode; (2) Both baselines perform worse than AdaVaR, indicating that, compared to simply training with diverse data, adaptive reasoning is more effective for developing general-purpose reasoning capabilities.

\begin{figure}[t]
  \centering
  \begin{subfigure}{0.38\linewidth}
  \centering
    \includegraphics[width=\linewidth]{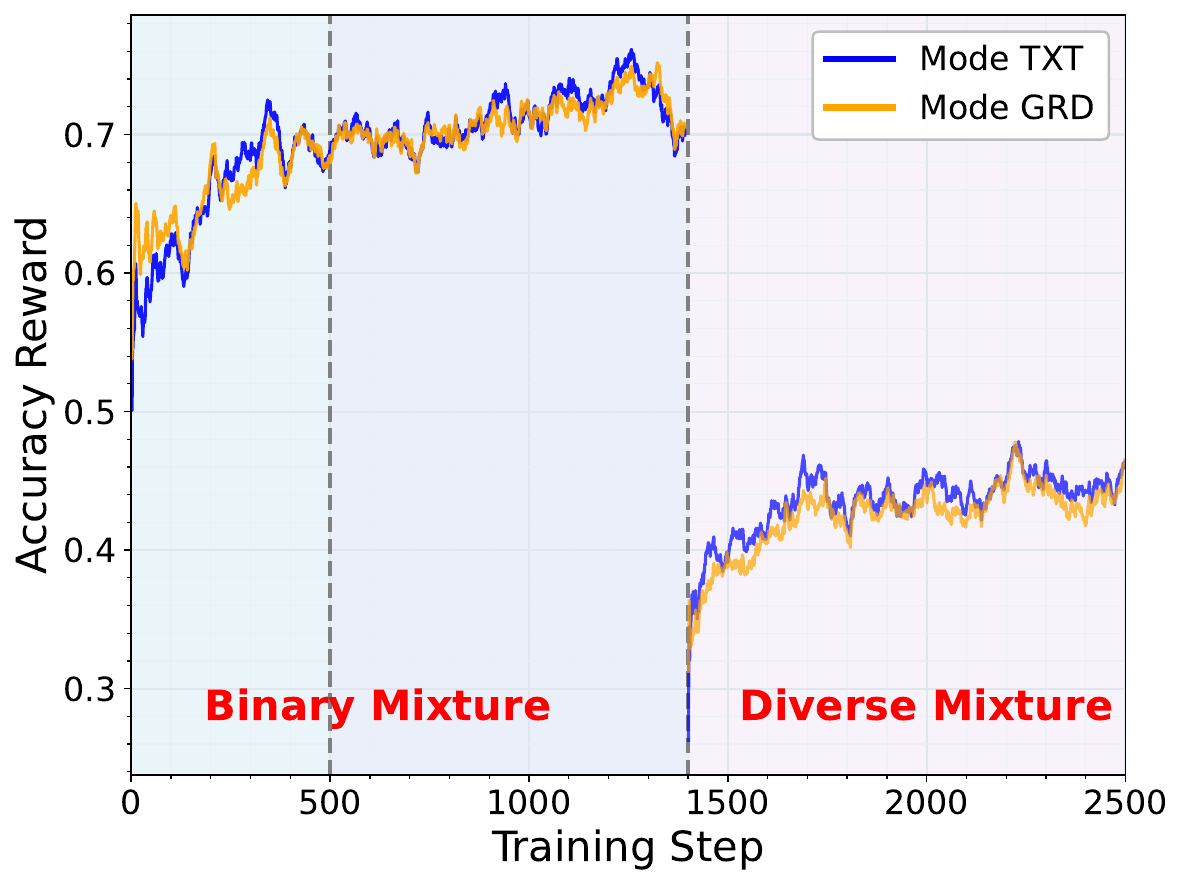}
    \vspace{-5mm}
    \caption{Training rewards on math problems.}
    \label{fig:curve-training}
  \end{subfigure}
  \hfill
  \begin{subfigure}{0.61\linewidth}
  \centering
    \includegraphics[width=\linewidth]{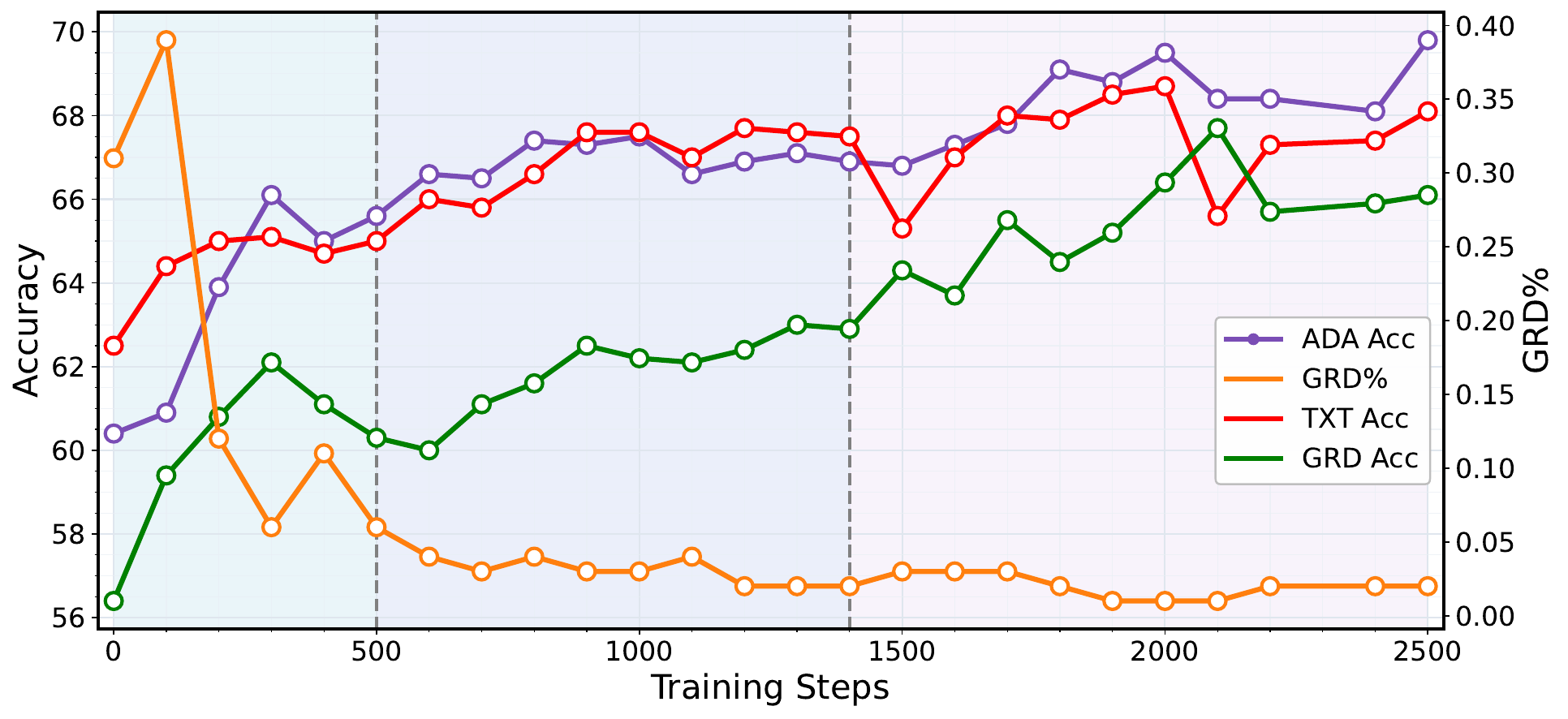}
    \vspace{-5mm}
    \caption{Performance evolution of different modes on MathVista.}
    \label{fig:curve-evaluation}
  \end{subfigure}
  \vspace{-5mm}
  \caption{Math-related training dynamics and evaluation metrics during Stage 2. ADA, TXT, and GRD are short for the three reasoning modes: adaptive, text-based, and grounded, respectively.}
  \label{fig:main-math-curve}
  \vspace{-3mm}
\end{figure}

\paragraph{Necessity of Mode-Specific Prefix} Another question is whether we need to distinguish between different modes. To investigate this, we remove the prefix token and mix the SFT data from two modes together, creating Mix-SFT-RL. Its performance is even inferior to the single-mode baselines. This suggests that simply mixing data from different modes is insufficient, and that the mode-specific prefix token introduced in AdaVaR helps the model differentiate between modes and facilitate uniform exploration across modes during the RL stage, resulting in a better overall performance.

\paragraph{The Upper Bound of Adaptive Reasoning} Table~\ref{table:adaptive_analysis} also presents the upper bound of adaptive reasoning: the performance when a prediction is considered correct if either mode yields the correct answer.
The upper bound clearly surpasses Qwen2.5-VL-3B and single-mode models. Even in mathematical problems, where text-based reasoning is more dominant, the grounded mode can solve some problems that the text-based mode cannot. 
This phenomenon demonstrates the complementary nature between two modes and highlights the great potential of the adaptive MoVT paradigm.
% Therefore, we believe the grounded reasoning mode introduces an effective inductive bias that complements the shortcomings of textual reasoning. Although the AdaVaR method helps the model approach its upper bound, there is still significant room for further exploration.

\input{tables/ablation}

\begin{figure}[t]
    \centering
    \includegraphics[width=\linewidth]{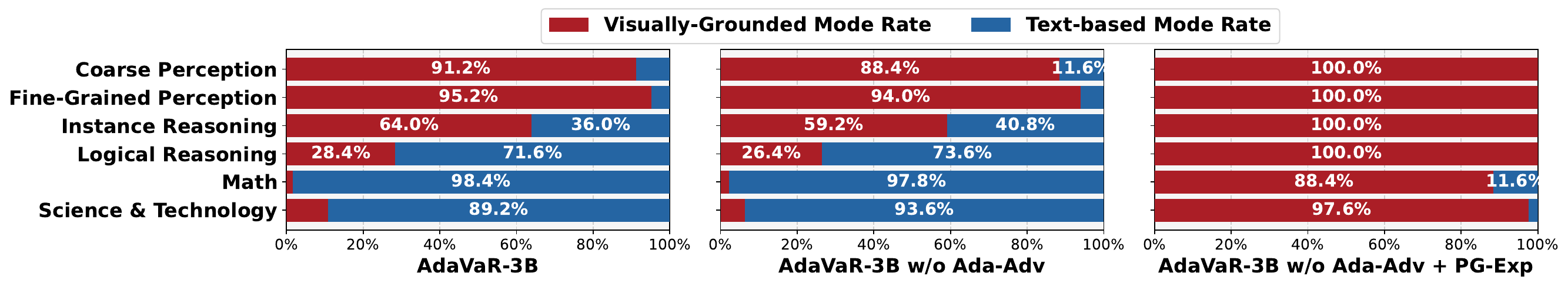}
    \vspace{-6mm}
    \caption{Proportion of each mode selected by different models across categories in MMStar.}
    \label{fig:grd_prop}
    \vspace{-2mm}
\end{figure}

\subsubsection{Question 2: Can AdaVaR select appropriate mode?}

Comparing the SFT and RL results in Table~\ref{table:adaptive_analysis}, we find that (1) the SFT model exhibits unexpected behavior: although the text mode performs significantly better on MathVista, it still chooses the grounded mode 31\% of the time, resulting in worse performance than the text mode alone. This also supports our argument in Section~\ref{section:intro}: it is hard to guide the model to learn effective mode selection during the SFT stage. (2) After Stage 2, AdaVaR exhibits reasonable mode selection: it prefers the text mode for math problems and the grounded mode for V* and POPE tasks (as well as fine-grained choices shown in Figure~\ref{fig:grd_prop}); its performance consistently surpasses single-mode reasoning across all benchmarks, indicating that the RL stage helps the model acquire effective mode selection abilities.
% (1) it is difficult to control the choice of modes in the SFT stage, AdaVaR-SFT exhibits unexpected selection—for example, choosing the grounded mode too often on MathVista and WeMath;
% (2) after AdaGRPO, AdaVaR acquires effective mode selection ability, most noticeably reflected in the fact that AdaVaR outperforms the single-mode results across all datasets.

\paragraph{Learning Curves of Mode Selection} How is the mode-selection capability acquired? Figure~\ref{fig:main-math-curve} illustrates the learning process using the mathematical scenario as an example.
%We find that the mode-selection preference is guided by the relative advantages across modes (reflected by the training rewards). 
It is observed that the relative performance across modes (as reflected by the relative reward levels of two modes in Figure~\ref{fig:curve-training}) guides the mode-selection preference, which is illustrated by the GRD\% curve in Figure~\ref{fig:curve-evaluation}.
The RL process can be divided into three phases, which are delineated in the figure using different background colors in Figure~\ref{fig:main-math-curve}:
(1) Exploration phase: at the beginning, performance estimates for different modes obtained via RL sampling are still unstable, and the GRD mode even outperforms the TXT mode for a period, with the choice of mode also fluctuating.
(2) Stabilization phase: as the text mode demonstrates a clear advantage in RL sampling, the adaptive model gradually tends to select the text mode. However, it does not yet achieve optimal performance, and the adaptive performance is slightly weaker than the TXT mode.
(3) Refinement phase: by introducing more difficult and diverse data, the capabilities of each mode steadily improve, and the model learns which types of math problems benefit from choosing the GRD mode, ultimately pushing the ADA mode to reach optimal performance.
Please refer to Appendix~\ref{appendix:training_dynamics} for the training dynamics in other domains, and to Appendix~\ref{appendix:mode_select_behavior} for a more comprehensive analysis of the learned mode selection behavior.

\paragraph{Effectiveness of AdaGRPO} Table~\ref{table:ablation_study} and Figure~\ref{fig:grd_prop} present the results of ablation experiments related to key designs in AdaGRPO. (1) When both Ada-Adv and PG-Exp are removed, which is equivalent to the original GRPO, we observe that RL reinforces the SFT model’s unreasonable preference for the grounded mode, resulting in overall inferior performance. (2) With PG-Exp included, the model is able to fully explores different modes and compares rollouts between them. In datasets where the advantages between modes are clear, such as V* and MathVision, the model makes reasonable mode selections and achieves good performance. (3) With further introduction of adaptive advantage, AdaVaR learns more refined mode selection abilities, leading to improvements on complex datasets like MathVista and MMStar, as well as further boosting overall performance.

\paragraph{Training Data Matters} Apart from demonstrating the importance of AdaGRPO, the results in Table~\ref{table:ablation_study} demonstrate that diverse data enables the model to learn more generalizable reasoning and mode selection abilities. Additionally, the curriculum learning strategy is also highly effective during the RL phase. Both factors contribute to the improved model performance.

% \subsubsection{Q3: How does AdaVaR learn the mode-selection ability?}

% \subsection{Further Analysis}

% \subsubsection{Effectiveness of the AdaVaR framework}

% Then we conduct ablation study to investigate the effect of components in the AdaVaR framework.

% \paragraph{Effectiveness of AdaGRPO}
% \subsubsection{Further Analysis}

\section{Conclusion}

This paper presents an adaptive visual reasoning paradigm, MoVT, for building general visual reasoning models. We propose AdaVaR, a framework that unifies the reasoning capabilities of different modes within a single model through SFT, and employ a tailored RL algorithm, AdaGRPO, to train the model to choose the appropriate mode based on the context. Experimental results demonstrate that, unlike existing models that excel only in specific domains, AdaVaR achieves consistent improvements across multiple scenarios--with AdaVaR-7B surpassing GPT-4o in average performance.

\section*{Ethics Statement}

As described in Appendix~\ref{appendix:train_data}, all data sources used in this paper are open-source and available for research purposes. We strictly comply with the corresponding terms of use to ensure our data can be used for scientific research. The original datasets have been widely used in related studies and are confirmed to contain no harmful content. Building on these datasets, we performed certain reconstruction and filtering steps, as detailed in Appendix~\ref{appendix:train_data}; this process only involves modifying the problem format and providing additional information for filtering. Through sampling-based spot checks, we have verified that this data construction process does not introduce harmful content, and our filtering is entirely for research purposes, without any discriminatory bias. 

Our reasoning models, AdaVaR-3B and AdaVaR-7B, currently focus on English scenarios, and we aim to expand to more languages in the future. Our work and artifacts are designed with the principle of universality and fairness, without any preference for specific demographic groups.

\section*{Reproducibility Statement}

We have taken rigorous steps to ensure that relevant researchers can easily reproduce this work. First, our approach is built upon the open-source Qwen2.5-VL-Instruct model. Second, in Section~\ref{section:method}, we describe our AdaVaR framework in detail, including how to represent reasoning generation sequences in a unified way (Section~\ref{section:mode_unification}), how to perform cold-start SFT (Section~\ref{section:SFT_stage}), and how to train the model’s mode selection capability via RL (Section~\ref{section:RL_stage}). Furthermore, we provide extensive additional details, including the implementation details and formulas of AdaGRPO (Section~\ref{section:adagrpo}), detailed model prompts (Appendix~\ref{appendix:detailed_prompts}), data construction details (Section~\ref{section:rl_data} and Appendix~\ref{appendix:train_data}), specific training parameter settings (Appendix~\ref{appendix:training_details}), and evaluation details (Appendix~\ref{appendix:evaluation_details}). In summary, we believe researchers can readily reproduce the findings of this paper. 
% During the review process, we provide reviewers with an anonymous code repository in \url{https://anonymous.4open.science/r/movt-12932}, including anonymized documentation, code, and sample data.
% After the anonymous review process, we will completely release 
Our data, model weights, and code are available at \url{https://github.com/Future-Living-Lab/mixture-of-visual-thoughts}.

\section*{Acknowledgment}
This work was supported by Alibaba Group through Alibaba Research Intern Program. The support was also provided by AI for Science Program, Shanghai Municipal Commission of Economy and Informatization (2025-GZL-RGZN-BTBX-02028), and National Key R\&D Program of China (2023YFF1204800). Computational resources were provided in part by CFFF.

\bibliography{iclr2026_conference}
\bibliographystyle{iclr2026_conference}

\appendix

\clearpage
\section{Appendix Roadmap}
To assist the reader in navigating our supplementary materials, we provide a brief overview of the Appendix below. Appendix~\ref{appendix:implementation_details} provides supplementary implementation details, including:
\begin{itemize}
    \item Detailed prompts in Appendix~\ref{appendix:detailed_prompts}.
    \item How to collect the training data (Appendix~\ref{appendix:train_data}).
    \item Training details in Appendix~\ref{appendix:training_details}.
    \item Details of the evaluation process in Appendix~\ref{appendix:evaluation_details}.
\end{itemize}

Appendix~\ref{appendix:supplement_analysis} presents supplementary experiments and analysis:
\begin{itemize}
    \item Appendix~\ref{appendix:mode_select_behavior} analyzes the mode-selection behavior learned by AdaVaR, including fine-grained selection, the impact of images on mode selection, and attention visualization.
    \item Appendix~\ref{appendix:adaptive_sft} explores whether the adaptive reasoning ability can be induced through SFT.
    \item A detailed error analysis is provided in Appendix~\ref{appendix:error_analysis}.
    \item Computational efficiency analysis is conducted in Appendix~\ref{appendix:computational_efficiency}.
    \item Appendix~\ref{appendix:training_dynamics} presents training dynamics during the two stages.
    \item The scalability of MoVT and AdaVaR, as well as the direction for future extension, are discussed in Appendix~\ref{appendix:future_direction}.
    \item Appendix~\ref{appendix:ablation} provides additional ablation studies on several designs.
    \item A comprehensive qualitative case study is included in Appendix~\ref{appendix:case_study}.
\end{itemize}

\section{Supplemented Implementation Details}
\label{appendix:implementation_details}

\subsection{Detailed Prompts}
\label{appendix:detailed_prompts}

Here we provide the complete prompt template used in our adaptive reasoning model:
\begin{center}
% \vspace{-0.2cm}
    \fcolorbox{black}{gray!10}{\parbox{.95\linewidth}{\textbf{System}: You are a helpful assistant. The user asks a question related to an image, you need to solve it. Please first think about the reasoning process in the mind and then provide the user with the answer. The reasoning process and answer are enclosed within \textless think\textgreater~ \textless/think\textgreater and \textless answer\textgreater~ \textless/answer\textgreater tags, respectively, i.e., \textless/think\textgreater reasoning process here \textless/think\textgreater ~\textless answer\textgreater answer here \textless/answer\textgreater. You have two modes of thinking, and you should choose the appropriate one based on the question: \\
    1. \textit{Grounded Reasoning}: Use this mode when you need to locate specific objects in the visual input. In your reasoning path, identify key objects and provide their corresponding bounding box coordinates in the format `object[x1, y1, x2, y2]'. When using this mode, begin your response with the tag \textcolor{blue}{\textless ground\textgreater}.\\
    2. \textit{Text-based Reasoning}: Use this mode for general reasoning based solely on textual thoughts. No object localization or coordinate output is required in this mode. When using this mode, begin your response with the tag \textcolor{blue}{\textless text\textgreater}. \\
    Choose the mode that best fits the task, and structure your response accordingly.\\
    \textbf{User}: \{QUESTION\}. Please first select the appropriate reasoning mode based on the question, using \textless ground\textgreater or \textless text\textgreater to indicate the type, then follow the corresponding format to output the reasoning process, and finally provide the correct answer according to the user requirement.\\
    \textbf{Assistant}: \textcolor{blue}{\textless mode prefix\textgreater} \textcolor{red}{\textless think\textgreater\;... \textless/think\textgreater\; \textless answer\textgreater\;... \textless/answer\textgreater}}}
    % \vspace{-0.2cm}
\end{center}
where the \{QUESTION\} will be replaced by the input question. Please note that we have added a post prompt after each question to emphasize mode selection, in order to prevent the model from ignoring the system prompt due to the long input sequence of images and questions.

\subsection{Details on Training Data Collection}
\label{appendix:train_data}

\subsubsection{SFT Data Collection}
\paragraph{Visually-Grounded Reasoning Data} Regarding the visually-grounded mode, we utilize the data constructed in VoCoT~\citep{li2025vocotunleashingvisuallygrounded}, we find that the dataset lacks multiple-choice samples, to avoid potential bias induced by such a distribution gap, we employ GPT-4o to rewrite a subset into multiple-choice format (by providing the original question and the correct option). To control the total size, we down-sample the GQA subset, resulting in a final dataset of 119K examples. 

During the re-formulation process, we strictly adhere to copyright regulations.
We would like to clarify that VoCoT data is released under the Apache 2.0 license, which permits free reconstruction, modification, and creation of derivative datasets. We strictly adhere to these requirements and will include the appropriate VoCoT attributions in our future open-sourcing process, clearly stating in our documentation that our dataset is derived from VoCoT with modifications.

For quality assessment, we sample 100 reconstructed samples to check for formatting errors or the generation of false-negative options. None of the samples exhibited any issues, which is primarily due to the simplicity of the reconstruction task and GPT-4o's strong instruction-following capability. Therefore, we believe the quality of the reconstructed data is reliable.

\paragraph{Text-based Reasoning Data} For the text-based mode, we aim to avoid introducing other inductive bias; therefore, instead of using an R1-OneVision–style approach~\citep{yang2025r1onevisionadvancinggeneralizedmultimodal} built from DeepSeek-R1, we distill Orsta~\citep{ma2025one}, a reasoning model obtained by applying RL directly to Qwen2.5-VL. The specific procedure is as follows: We randomly sample a total of 250K questions from R1-OneVision~\citep{yang2025r1onevisionadvancinggeneralizedmultimodal}, llava-CoT~\citep{xu2024llava}, NuminaMath1.5~\citep{numina_math_datasets}, and Virgo~\citep{du2025virgo}. For each question, Orsta generates eight responses with temperature of 1. We then use a rule-based method to determine correctness and retain only the correct instances; if multiple reasoning traces are correct, we randomly select one during training. We also observe that Orsta might skip the reasoning process and directly output an option, so we distinguish reasoning data from direct-answer data based on whether the model’s reply contains the \textless think\textgreater~and \textless answer\textgreater~tags, and apply different system prompts for these two data types. After the rejection sampling, we obtain 115K reasoning examples and 95K direct-answer examples.
% As for the text-based data, we utilize 

\subsubsection{RL Data Collection}

\paragraph{Difficulty-based Filtering} Since overly simple questions cannot provide effective supervision during the RL process, we filter them out from the RL data. Following prior work~\citep{ma2025one,zheng2025deepeyesincentivizingthinkingimages}, we adopt a model-based evaluation strategy: questions that are answered correctly in all 8 random trials by Qwen2.5-VL-7B are considered too simple and are excluded.

\paragraph{RL Data Collection} As introduced in Section~\ref{section:rl_data}, our RL dataset consists of two parts: (1) existing data with verifiable answers, including math-related Geo170K~\citep{gao2023gllava} and MM-Eureka~\citep{meng2025mmeurekaexploringfrontiersmultimodal}, as well as OmniCount~\citep{mondal2025omnicount} for object counting. We further filter MM-Eureka following the process: we first perform uniform sampling across the data sources (capping any category exceeding 1,500 instances at 1,500) and remove overly simple questions using the difficulty-based filtering method described in the previous paragraph, ultimately retaining 13K examples. (2) Additionally constructed data for general scenarios.
% source of the data
The questions and answers come from the SFT data used in LLaVA-OneVision~\citep{li2024llavaonevisioneasyvisualtask} and InternVL~\citep{zhu2025internvl3}. We retain only the subsets that are related to mathematics, OCR, object counting, STEM, grounding, and table/document understanding.
% post process (category)
For this part of data, we use an offline Qwen2.5-VL-72B~\citep{bai2025qwen25vltechnicalreport} model to assign task categories to each samples, using the same categorization scheme as in Orsta~\citep{ma2025one}.
% post process (answer verifiable)
% post process (difficulty)

To ensure that answers could be verified by a rule-based reward function, we first use Qwen2.5-VL-72B to determine the answer type. Depending on the task and question type, we collect and evaluate responses from multiple models via offline experiments, and, guided by the question type, we augment some input queries with prompts specifying the required answer type. Ultimately, we retain only those questions that could be judged accordingly.
Finally, after filtering by difficulty to remove overly easy questions, we retain 22K examples.

\paragraph{Multi-Modal Data Processing} Since our model is based on Qwen2.5-VL~\citep{bai2025qwen25vltechnicalreport}, our data processing methods strictly follow those of Qwen2.5-VL: we apply the same dynamic resolution strategy for patchifying images and use the corresponding tokenizer for tokenizing text.

\begin{table*}[t]
    \centering
    \caption{The detailed training hyper-parameters of AdaVaR.}
    \vspace{-2mm}
    \resizebox{0.94\linewidth}{!}{
        \begin{tabular}{lcc}
        \toprule
        Configuration & Stage 1 SFT & Stage 2 RL \\ 
        \midrule
        Model Initialization  & Qwen2.5-VL-Instruct & Stage1  \\
        Trainable Modules & MLP + LLM & All Modules \\
        Optimizer  & AdamW & AdamW \\
        Optimizer Hyperparameters & $\beta_1=0.9$, $\beta_2=0.95$, $\epsilon=1e^{-6}$ & $\beta_1=0.9$, $\beta_2=0.95$, $\epsilon=1e^{-6}$ \\
        Global batch size & 256 & 32 \\
        Peak learning rate & 5e-6 & 1e-6 \\
        Learning rate scheduler & Cosine & Linear \\
        Training Epochs & 1 & 1 \\
        Warm-up ratio & 0.01 & 0.0 \\
        Weight decay & 0.1 & 0.0 \\
        Max Sequence Length (SFT) & 4096 & - \\
        Max Prompt Length (RL) & - & 2048 \\
        Max RL Completion Length (RL) & - & 2048 \\
        KL Penalty Coefficient (RL) & - & 0.04 \\
        Generation Temperature (RL) & - & 0.9 \\
        \# Rollouts per Sample (RL) & - & 8 \\
        Numerical precision & bfloat16 & bfloat16 \\
        GPU Usage & 8 NVIDIA A100 & 32 NVIDIA A100 \\
        DeepSpeed Configure & zero3  & zero3 \\
        Training Time & 4h & 29h \\
        \bottomrule
        \end{tabular}
    }
    \label{table:hyper}
    \vspace{-4mm}
\end{table*}

\paragraph{Summary on Data Efficiency} Our AdaVaR framework operates under a data-efficient setting, in terms of (1) Data sources: We rely entirely on existing datasets, and the processes of data reconstruction and filtering do not involve any additional human annotation.
(2) Data volume: We use 329K samples in the SFT phase and 55K in the RL phase, totaling 384K—comparable to other RL methods such as OVR-7B~\citep{wei2025open} (428K), and significantly less than the SFT data used for mainstream LVLMs, such as 1.8M samples used in LLaVA-OneVision~\citep{li2024llavaonevisioneasyvisualtask}.

\subsection{Training Details}
\label{appendix:training_details}

We list the hyper-parameter settings in Table~\ref{table:hyper}, and here we discuss the rationale behind several key parameter choices.

\paragraph{The Number of Training Epochs in the SFT Stage} We follow the commonly adopted setup in existing works~\citep{yang2025r1onevisionadvancinggeneralizedmultimodal,chen2025sft} to train one epoch for SFT. We validate the rational behind this setting in Appendix~\ref{appendix:training_dynamics}.

% We also present the loss curve of the SFT stage in Figure~\ref{fig:sft-loss}. As illustrated, the loss has already converged within one epoch; therefore, to prevent the model from overfitting to the SFT data, we stop after one epoch.

\paragraph{The Number of Rollouts Generated per Sample} In the RL stage, ``\# Rollouts per Sample'' refers to the number of rollouts generated for each question. Generating more rollouts per sample can help provide better RL supervision signals and improve performance, but it also incurs higher computational costs. Therefore, to ensure a fair comparison with other models in Table 1, we set it to 8, which corresponds to $n=4$ in Figure~\ref{fig:AdaGRPO} and the equations in Section~\ref{section:RL_stage}, meaning that the policy model generates 4 rollouts with each reasoning mode, for a total of 8. Similarly, for a fair comparison with ablated models, when training the ablated models that are based on a single reasoning mode in Section~\ref{section:delving}, we set the number of rollouts generated in GRPO to 8. 

\paragraph{Coefficient for KL Penalty} For brevity, the KL penalty are omitted in equations in Section~\ref{section:RL_stage}. In practice, we introduce the KL penalty term to preserve the knowledge of reasoning modes learned during the SFT phase. The coefficient is set to 0.04, following the settings used in existing works and frameworks~\citep{chen2025r1v,shen2025vlm,openr1}.
% we include the KL penalty item in our RL training loss with the coefficient of 0.04. 

\paragraph{Training Dynamics Monitor} During training, we implement a monitor to track the performance of different modes on various types of data, allowing us to better observe the relative strengths and weaknesses between modes, the detailed training dynamics are provided in Appendix~\ref{appendix:training_dynamics}.

\begin{algorithm}[t]
\caption{Mode-relative advantage calculation.}\label{alg:mode_adaptive}
\begin{algorithmic}[1]
\Require  $r[1..2n]$ \textcolor{blue}{\# rewards of all rollouts where $r[1:n]$ and $r[n+1:2n]$ correspond to the text-based mode and the visually-grounded mode, respectively.}
\Require  $\Phi(x)$ \textcolor{blue}{\# CDF of the standard Gaussian distribution $N(0,1)$}
\Ensure $A_t,A_v$
\State \textcolor{blue}{\# mean and standard deviation of text-based mode}
\State $\mu_t\gets \text{mean}(r[1:n])$, $\sigma_t\gets \text{std}(r[1:n])$
\State \textcolor{blue}{\# mean and standard deviation of visually-grounded mode}
\State $\mu_v\gets \text{mean}(r[n+1:2n])$, $\sigma_v\gets \text{std}(r[n+1:2n])$
\State \textcolor{blue}{\# Advantage of text-based mode over visually-grounded mode}
\State $A_t=\Phi(\frac{\mu_t-\mu_v}{\sqrt{\sigma_v^2+\sigma_t^2}})$
\State \textcolor{blue}{\# Advantage of visually-grounded mode over text-based mode}
\State $A_v=1-A_t$
\State \textbf{return} $A_t,A_v$
\end{algorithmic}
\end{algorithm}

\paragraph{Algorithm for Advantage Calculation} Specifically, in Algorithm~\ref{alg:mode_adaptive}, we provide the pseudo code for calculating the mode-relative advantage $A_v$ and $A_t$ in Section~\ref{section:adagrpo}.

\subsection{Evaluation Details}
\label{appendix:evaluation_details}

\paragraph{Benchmark Details} As indicated in the header of Table~\ref{table:main_results}, we use the testmini split in MathVista~\citep{lu2023mathvista}, the test split in MathVision~\citep{wang2024measuringmultimodalmathematicalreasoning}, the vision-only portion of the testmini split in MathVerse~\citep{zhang2024mathversedoesmultimodalllm}, the val split in MMStar~\citep{chen2024we}, the test split in V*~\citep{wu2023textit}, the testmini split in WeMath~\citep{qiao2024we}, all four splits of POPE~\citep{li2023evaluating}, and the hard split of SpatialScore~\citep{wu2025spatialscore}. For the eight benchmark datasets used in this paper, the evaluated model is required to produce structured outputs (e.g., option letters or numbers), we prompt the model to generate the desired format, and evaluate the results offline using rule-based procedures. Except for WeMath, for which we report the strict accuracy as specified in the original paper~\citep{qiao2024we}.

\paragraph{Average Schema} Since each benchmark represents a different domain, but the number of samples in each is inconsistent, all benchmarks are weighted equally for the average accuracy. We aim for the average performance to comprehensively reflect capabilities across multiple domains.
% —i.e., generalization ability. Averaging based on the number of benchmarks might lead to bias toward specific benchmarks the average accuracy across all questions is reported for other benchmarks.

\paragraph{Mode-Switching Mechanism} Specifically, the phrase ``getting stuck in repetitive logic" mentioned in Section~\ref{section:experimental_setup} refers to the situation where the mode reaches the maximum number of tokens (8192, significantly longer than the typical reasoning length of AdaVaR) without generating an answer in the form of \textless answer\textgreater \textless/answer\textgreater. Please note that our mode-switching mechanism does not leverage any additional information, ensuring a fair comparison with other models; rather, mode switching is a new feature enabled by our MoVT reasoning paradigm.

\paragraph{Compared Models} As discussed in Section~\ref{section:intro}, the compared models can be divided into two groups: text-based models including LMM-R1~\citep{peng2025lmm}, VLAA-Thinker~\citep{chen2025sft}, MM-Eureka~\citep{meng2025mmeurekaexploringfrontiersmultimodal}, OVR-7B~\citep{wei2025open}, and Orsta-7B~\citep{ma2025one}; as well as the visually-grounded models including GRIT~\citep{fan2025gritteachingmllmsthink}, ViGoRL~\citep{sarch2025grounded}, DeepEyes~\citep{zheng2025deepeyesincentivizingthinkingimages}, and Chain-of-Focus~\citep{zhang2025chainoffocusadaptivevisualsearch}. 
Since most compared models focus on specific domains and do not provide evaluation results across all benchmarks, we conduct evaluation to fill in the missing results for a comprehensive comparison. Specifically, to ensure a fair comparison, we do not retrain the corresponding models; instead, we strictly follow the original papers, using the officially released pretrained checkpoints and evaluating them with the recommended parameters according to their respective evaluation protocols.
% As shown in Table~\ref{table:main_results}, most models are specialized in particular domains. Therefore, under a fair, standardized experimental setting, we provide additional evaluation results for multiple models to enable cross-benchmark average-performance comparisons.

\paragraph{Evaluation Environment and Cost} All evaluation is conducted on a single node with 8 A100 GPUs, accelerated by the vLLM framework~\citep{kwon2023efficient}. As measured in Appendix~\ref{appendix:infer_efficiency}, a complete evaluation of the 3B-scale model across all benchmarks requires 17.7 GPU hours (on A100), making frequent performance monitoring during evaluation highly resource-intensive. Therefore, we only track MathVista performance, as shown in Figure~\ref{fig:main-math-curve}.

\subsection{LLM Usage}

Our use of LLMs in this paper mainly comprises two parts: writing assistance and tools for data construction. During writing, we use GPT-4o to help polish the exposition. In the data construction process, as described in Appendix~\ref{appendix:train_data}, we employ Qwen2.5-VL and GPT-4o to assist with data reconstruction, annotation, and filtering.

\section{Supplementary Experimental Results and Analysis}
\label{appendix:supplement_analysis}

\subsection{Delving into Mode-Selection Behaviors}
\label{appendix:mode_select_behavior}

In the main text, we have discussed mode selection at the dataset level and the coarse-grained task level within MMStar. This section provides additional analysis to complement that discussion.

\subsubsection{Mode Selection at the Fine-Grained Task Level
}
\label{appendix:more_fine_grained_selection}

\begin{figure}[t]
    \centering
    \includegraphics[width=\linewidth]{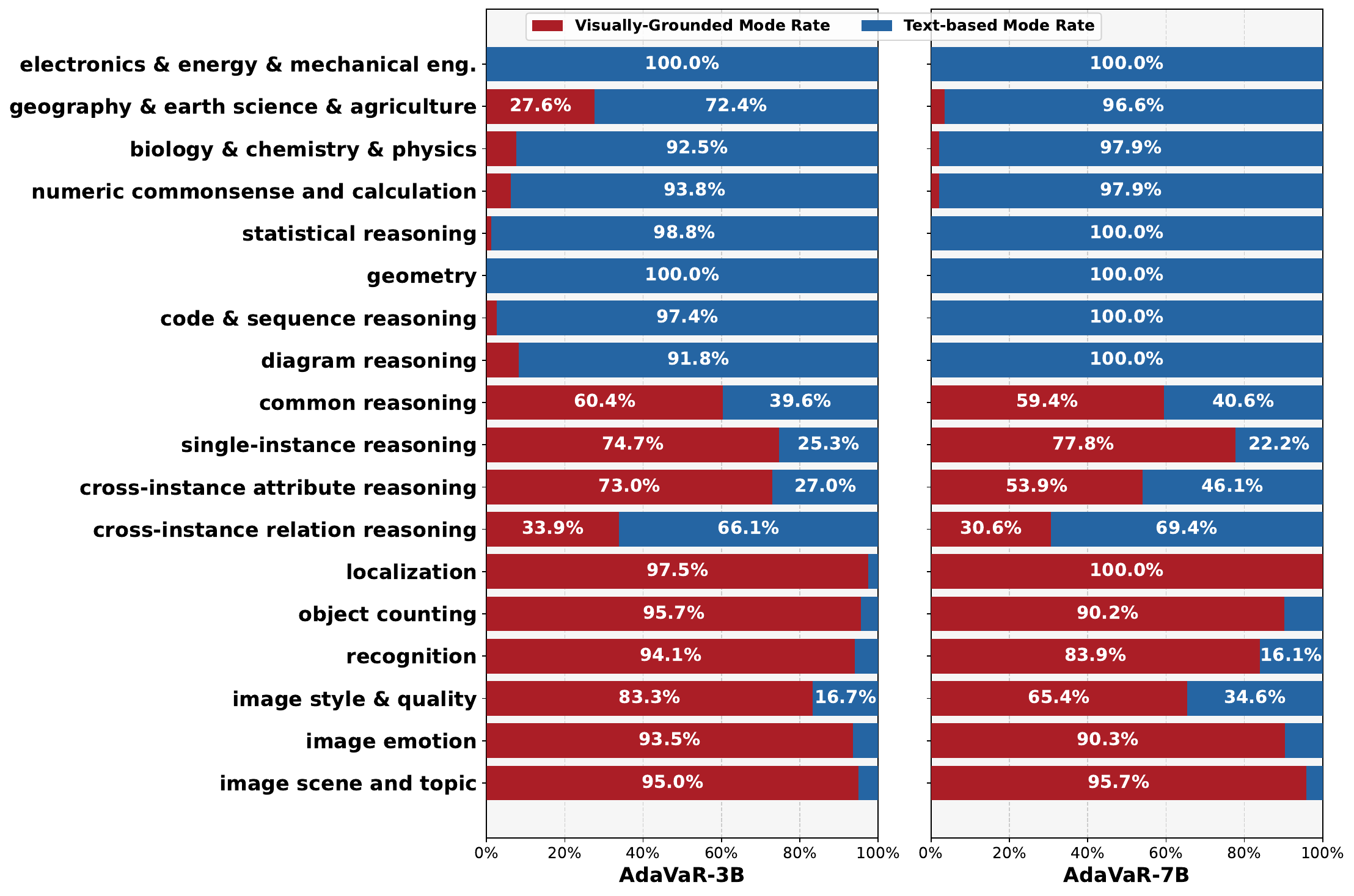}
    \vspace{-4mm}
    \caption{Proportion of each mode selected by different models across sub-categories in MMStar.}
    \label{fig:fine-grained-grd-prop}
    \vspace{-2mm}
\end{figure}

Figure~\ref{fig:fine-grained-grd-prop} illustrate the fine-grained mode selection tendencies of AdaVaR-3B and AdaVaR-7B. We observe that, since the 7B model has richer knowledge, AdaVaR-7B prefers the text-based mode for knowledge-related categories, such as geography \& earth science \& agriculture. This also highlights the necessity of the RL process: we need to sample during RL to better estimate the strengths and weaknesses of each mode, which are different for different models (there is no universal ground truth of mode selection that is applicable for all models).

\input{tables/mode_selection_image_types}

\subsubsection{The Impact of Image Types on Mode Selection}

Furthermore, we conduct an analysis from the perspective of images. We employ GPT-5 to annotate the image types in MMStar and analyze their impact on mode selection. As presented in Table~\ref{table:mode_select_image_types}, image type also has a significant impact on mode selection:
(i) for natural images, the model tends to use the grounded mode to precisely locate and analyze key objects;
(ii) when images contain more abstract or harder-to-localize concepts—such as artistic styles in artwork, content in documents and diagrams, or angles in geometry problems—the model tends to favor the text-based reasoning mode.

\subsubsection{Attention Visualization for Mode Selection}

To better understand how MoVT selects between the text-based and grounded modes, we visualize the attention patterns used when predicting the mode prefix. Concretely, for each sample we probe the attention scores from the token immediately preceding the mode prefix (i.e., the token that ``decides'' the mode) to all previous tokens. Following~\citep{kaduri2025s}, we select the middle 25\% of layers and average the attention weights across these layers and across all attention heads. 

\begin{figure}[t]
    \centering
    \includegraphics[width=0.92\linewidth]{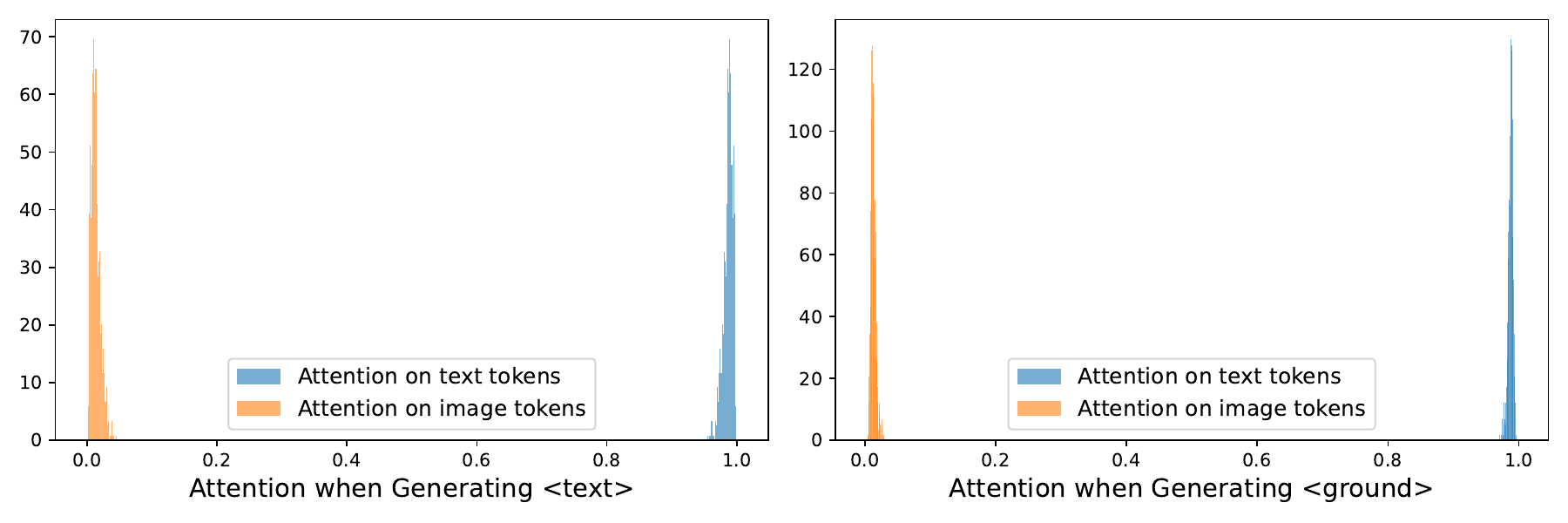}
    \caption{Distribution of attention mass over text tokens (blue) and image tokens (orange) when generating the \texttt{<text>} (left) and \texttt{<ground>} (right) mode prefixes on MMStar. }
    \label{fig:mmstar_attn_text_image}
\end{figure}

We first aggregate the attention mass over all image tokens and all text tokens on the MMStar dataset. As shown in Figure~\ref{fig:mmstar_attn_text_image}, when generating either the \texttt{<text>} or \texttt{<ground>} prefix, the model places the vast majority of attention on text tokens (0.97 in average) rather than image tokens (0.03 in average). This is consistent with the nature of MMStar, where the question type is primarily encoded in the textual prompt and multiple questions can be asked about the same image.

\begin{figure}[t]
    \centering
    \includegraphics[width=0.92\linewidth]{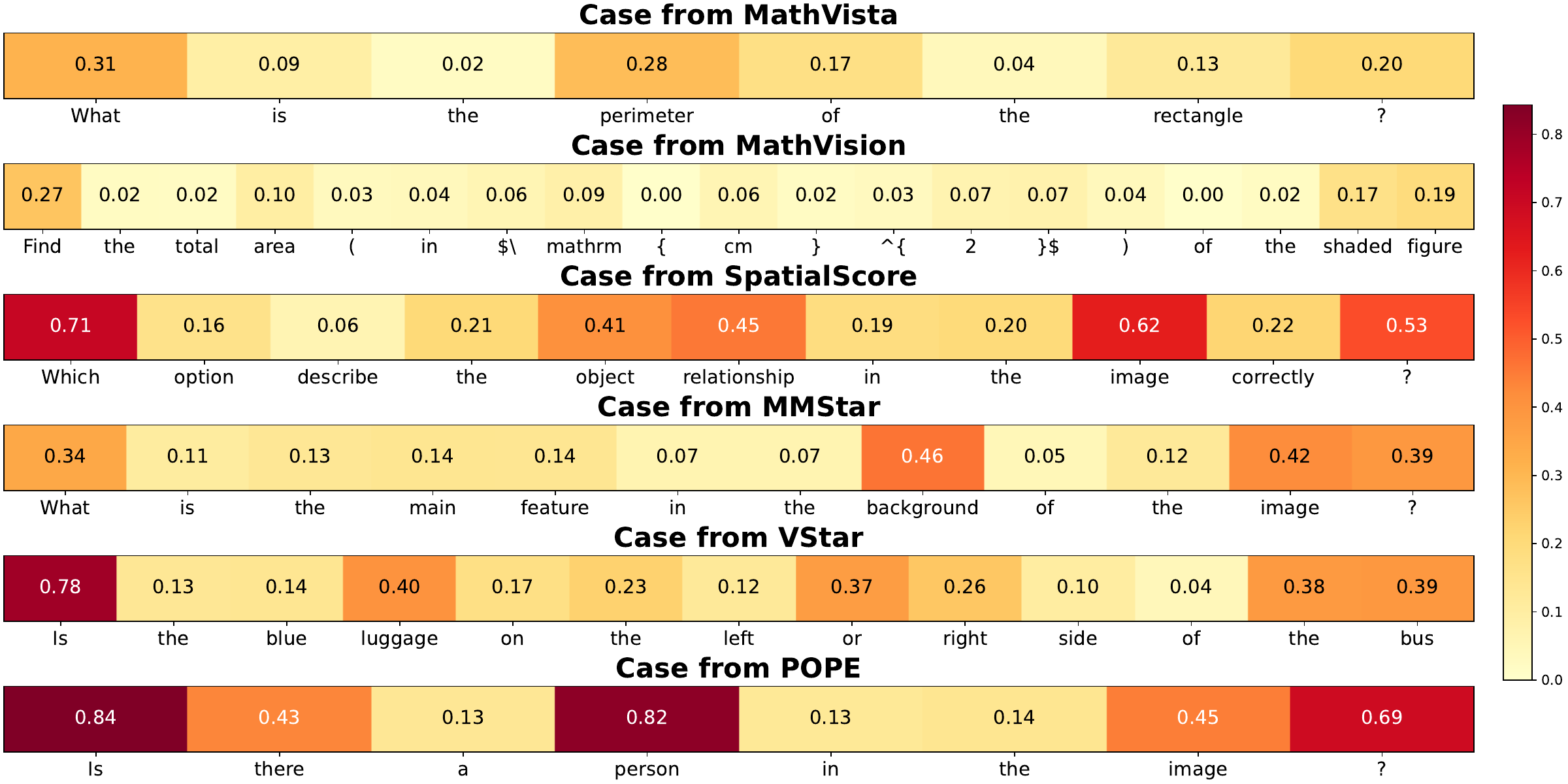}
    \caption{Token-level attention heatmaps over the question when generating the mode prefix.}
    \label{fig:attn_token_heatmap}
\end{figure}

We further examine how attention is distributed over different textual tokens. Figure~\ref{fig:attn_token_heatmap} visualizes token-level attention maps for representative examples from six benchmarks. We observe that the attention concentrates on tokens that define the question type, such as the interrogative words ``which'', ``what''; key mathematical terms like ``
perimeter'', ``area'' and object-oriented concepts of ``background'', ``bus'' and ``person''. 

\input{tables/naive_ensemble}

\subsubsection{Comparison with Naive Mode Ensemble Strategies}

To validate the effectiveness of the mode selection learned by AdaVaR, we compare it with two naive strategies to use external knowledge for mode ensemble at the dataset level: (1) Expert pick: selection based on expert knowledge—here, we choose the text-based mode for math benchmarks and the grounded mode for others. Note that, in practical applications, even experts would find it difficult to make reasonable judgments on mixed-task datasets like MMStar and SpatialScore.
(2) Max(GRD, TXT): for each dataset, we select the mode that achieves higher accuracy. Note that this strategy uses additional information from the dataset, making it inherently unfair for comparison.

The results in Table~\ref{table:naive_ensemble} show that both strategies—even with additional information—are outperformed by the selection made by AdaVaR. This indicates that: (i) mode selection at the individual sample level is necessary, especially for diverse queries in datasets like MMStar and in real-world applications; and (ii) our model is capable of making highly reasonable and effective mode choices.

\input{tables/supplement_table}

\subsection{Can the ability to adaptively select modes be learned via SFT?}
\label{appendix:adaptive_sft}

To further validate the claim made in Section~\ref{section:intro}—that it is difficult to control the model’s ability to choose among different modes during the SFT stage—and to demonstrate the necessity of the RL stage, we conduct an supplementary experiment. Without relying on RL at all, we train the model using two stages of SFT. The first stage is identical to AdaVaR Stage 1 and is used to learn the reasoning capabilities of different modes; the second stage is based on sampling from the model trained in Stage 1 to offline estimate the relative performance of the different modes.

Specifically, using the same RL data as AdaVaR, we estimate for each question with AdaVaR-SFT by running inference eight times with each of the two modes (temperature=1), estimating the accuracy of both modes, and retaining the rollouts of the mode with higher accuracy; if the accuracies are equal, we keep both. We then aggregate the reasoning trajectories corresponding to the estimated better mode(s) across all questions to serve as the SFT data for the second stage.

As presented in Table~\ref{table:sft_adaptive_analysis}, SFT-Adaptive-3B indicates the SFT-guided adaptive reasoning baseline.
% It can be observed that SFT-Adaptive-3B shows a clear preference for the grounded mode, which is unwarranted based on the evaluation results, as its overall performance is even worse than the AdaVaR-SFT-3B model after Stage 1 (except). A possible reason is that, as shown in Figure~\ref{fig:curve-training}, after SFT the performance of the two modes has not yet converged to a stable level (the text-based mode still has substantial room for improvement). Due to the limitations of the SFT framework, we can only perform offline sampling, which further reinforces this preference, as reflected in SFT-Adaptive-3B.
It can be seen that SFT-Adaptive-3B exhibits a clear preference for the grounded mode; according to the evaluation results, this is unwarranted, because its performance on mathematical tasks and its overall performance are even worse than the model after Stage 1 (except on POPE and VStar, which are dominated by the grounded mode). A possible reason is that, as shown in Figure~\ref{fig:curve-training}, after SFT the performances of the two modes have not yet stabilized (the text mode still has substantial room for improvement). Due to the limitations of the SFT framework, we can only perform offline sampling, which further reinforces this preference, as reflected in SFT-Adaptive-3B.

\input{tables/response_type}
\input{tables/error2question_type}
\input{tables/error2image_type}
\input{tables/train_efficiency}

% Overall, using SFT to guide the model’s adaptive selection can be regarded as a form of offline reinforcement learning; within a single iteration it shows no clear efficiency advantage and may even require multiple iterations. In contrast, RL operates in an online learning setting and has noticeably higher training efficiency. In summary, RL is a more suitable learning framework for adaptive reasoning. 
Overall, as an offline learning framework, SFT finds it is difficult to both strengthen different modes and guide mode selection, and it cannot correct unreasonable biases that emerge during iteration (and may even exacerbate them). In contrast, RL, as an online learning framework, uses dynamic sampling and estimation to effectively help the model enhance its reasoning ability and assess the relative merits of the modes. Therefore, the RL stage in the AdaVaR framework is necessary.
However, as a promising future direction, it is worth exploring a multi-round SFT+RL iterative framework to further improve the model’s capabilities.

\input{tables/inference_efficiency}

\subsection{Error Analysis}
\label{appendix:error_analysis}

First, we categorize errors into two types: (i) mode selection error (SE): the model fails to select the mode that could have produced the correct answer; (ii) reasoning error (RE): neither mode is able to produce the correct answer. As shown in Table~\ref{table:response_type}, across all datasets, mode selection errors constitute only a small portion of the total errors (all less than 1/4).

Furthermore, we investigate the distribution of mode selection errors across different question types and image types in MMStar. Results in Table~\ref{table:error2question_type} and Table~\ref{table:error2image_type} show that: (1) most errors occur in domains where one mode has a clear advantage, such as perception tasks and natural images, where the grounded mode performs better, and science or math problems, where the text-based mode holds the advantage. (2) These exceptions illustrate that the complementarity between the two modes exists not only at the domain level, but also at the sample level. As a result, an optimal mode selection strategy cannot be predetermined and must be learned through the AdaGRPO algorithm.

Overall, our results indicate: (i) the context-adaptive mode selection in MoVT is reasonable, and (ii) our AdaVaR framework is capable of learning effective selection strategies, (iii) while still leaving room for further optimization.

\subsection{Analysis of the Computational Efficiency}
\label{appendix:computational_efficiency}

\subsubsection{Training Efficiency: GRPO versus AdaGRPO}
\label{appendix:train_efficiency}

First, as stated in Section~\ref{section:adagrpo} and Appendix~\ref{appendix:training_details}, to ensure a fair comparison between GRPO, AdaGRPO generates $n=4$ rollouts per mode (8 per sample), while GRPO generates $2n=8$ rollouts per sample. In this section, we verify the training efficiency both theoretically and empirically.

From a theoretical perspective, the main computational cost is incurred during rollout generation and the calculation of the training objective: (i) In rollout generation, simply generating rollouts separately for each mode would require two KV cache computations, leading to additional cost. However, in our design, the two modes share the same input sequence except for the mode-specific prefixes. By sharing the KV cache across modes, we achieve nearly the same computational cost as GRPO. (ii) Regarding the training objective, both algorithms require forward passes of the policy model and reference model on 8 rollouts, so their computational costs are also nearly identical.

For empirical validation, we implement a FLOPS tracker using the deepspeed library and monitor the wall-clock time during training. As presented in Table~\ref{table:training_efficiency}, both theoretical computation and actual execution statistics show that the efficiency of AdaGRPO and GRPO is nearly identical (with approximately 4\% additional overhead).

\subsubsection{Inference Efficiency}
\label{appendix:infer_efficiency}

As for inference, theoretically, as described in Section~\ref{section:mode_unification}, our AdaVaR model naturally integrates mode selection and reasoning into a single sequence generation step. During inference, unlike the training phase, there is no need to estimate mode-wise advantages. Therefore, similar to single-mode reasoning models, AdaVaR only requires a single call to the ``model.generate()'' method for each sample. Empirically, we compare the inference efficiency of AdaVaR and single-mode baselines in Table~\ref{table:inference_efficiency}. As illustrated, the inference efficiency of different reasoning models is very similar on average, demonstrating that our framework does not introduce additional inference cost.

\subsection{Training Dynamics}
\label{appendix:training_dynamics}

\begin{figure}[t]
  \centering
  \begin{subfigure}{0.525\linewidth}
  \centering
    \includegraphics[width=\linewidth]{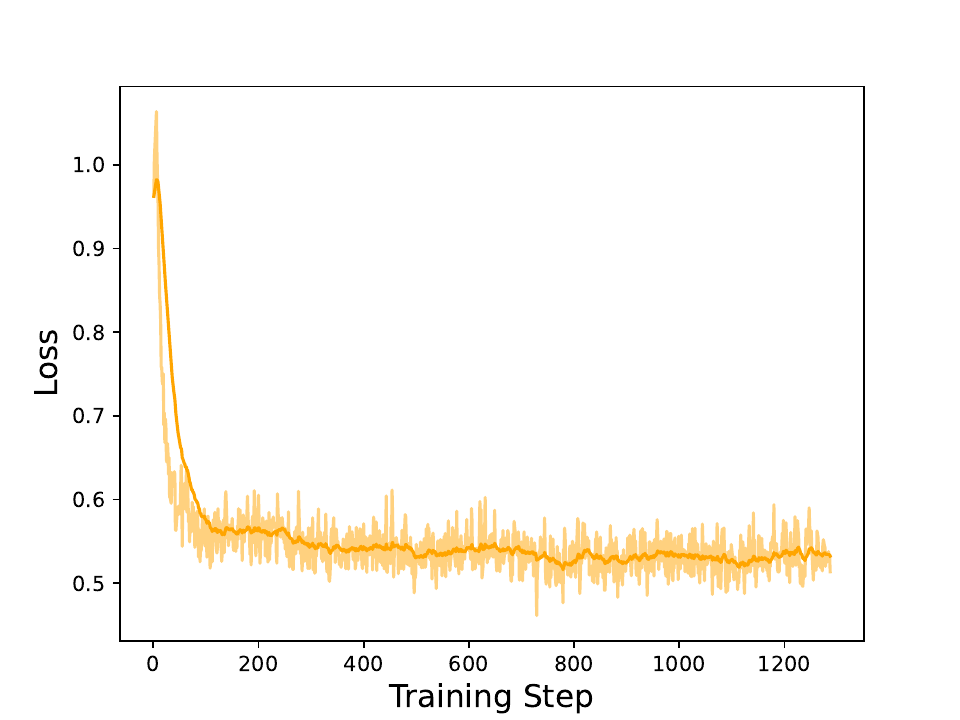}
    \caption{Training loss in the SFT stage.}
    \label{fig:sft-loss}
  \end{subfigure}
  \hfill
  \begin{subfigure}{0.465\linewidth}
  \centering
    \includegraphics[width=\linewidth]{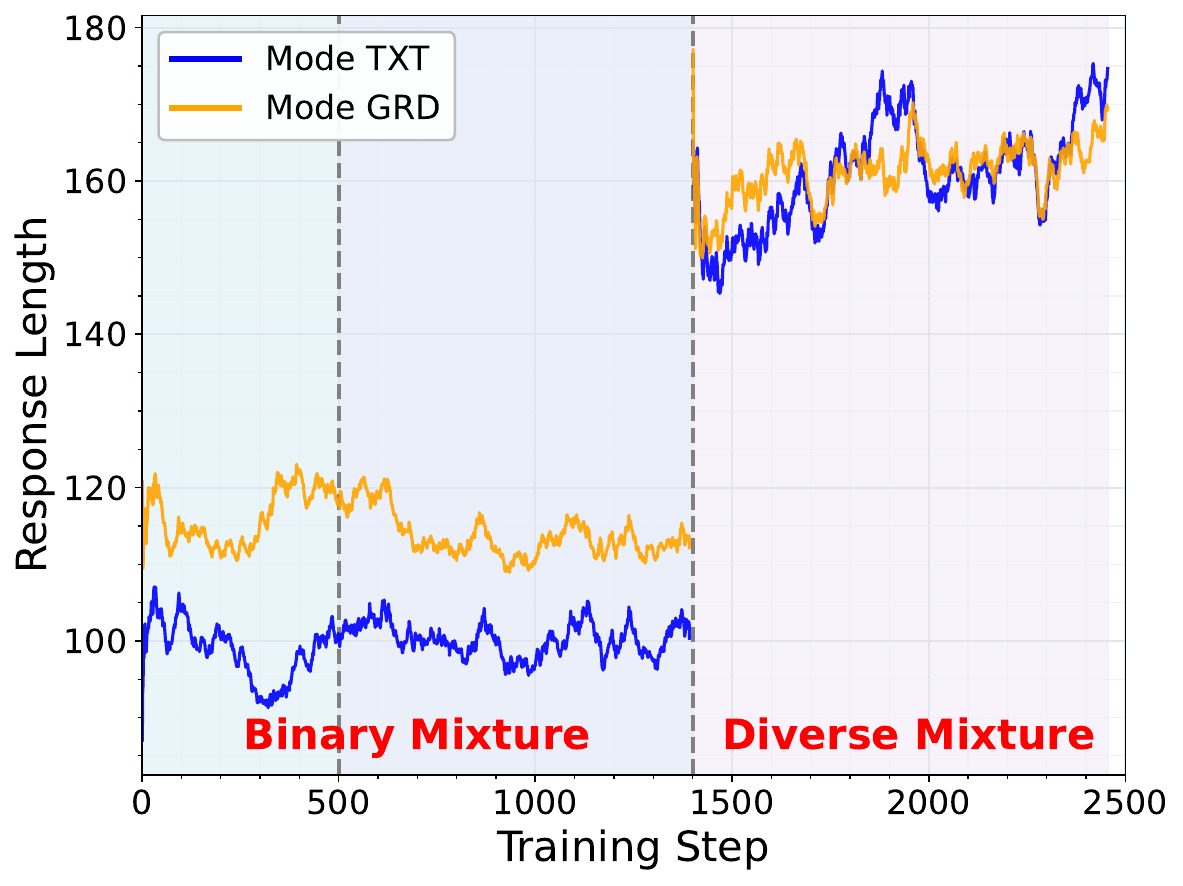}
    \caption{Response lengths during the RL stage.}
    \label{fig:rl-length}
  \end{subfigure}
  \vspace{-5mm}
  \caption{Supplementary training curves.}
  \label{fig:supplement-curve}
\end{figure}

\paragraph{Loss Curve in the SFT Stage} Figure~\ref{fig:sft-loss} shows the training loss during the SFT stage, demonstrating that the training process has already converged after one epoch. 

\begin{figure}[t]
    \centering
    \includegraphics[width=0.98\linewidth]{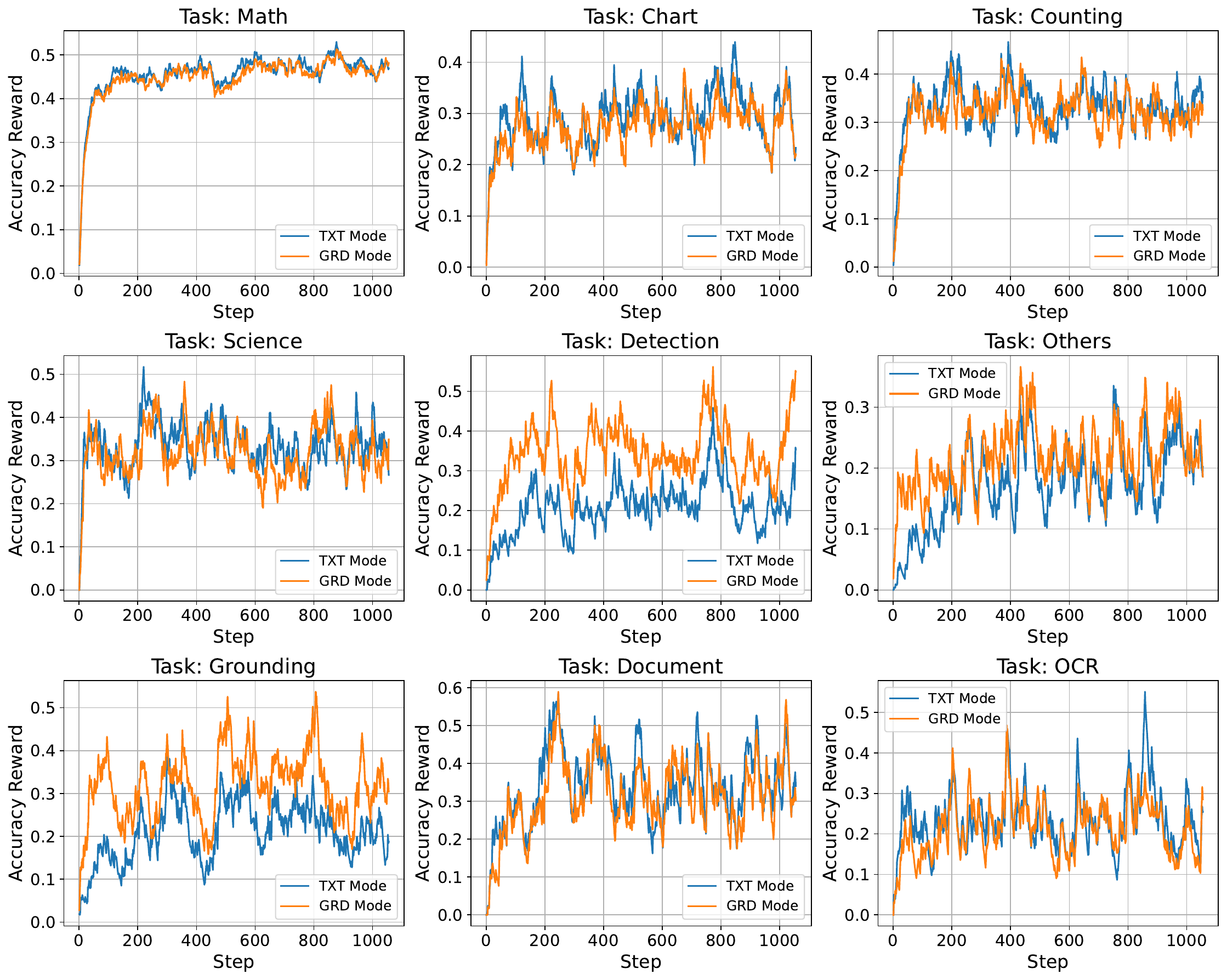}
    \vspace{-2mm}
    \caption{Accuracy reward curves for different modes across different tasks during RL.}
    \label{fig:training_dynamics}
    \vspace{-2mm}
\end{figure}

\paragraph{Rewards in the RL Stage} Figure~\ref{fig:training_dynamics} shows the training dynamic of the two reasoning modes across different task during Stage 2 (the diverse mixture), illustrating the relative strengths and weakness between two modes. 
Based on the comparative performance of the two modes, we find that the text-based mode predominates on Math and Science, while the grounded mode excels on Detection and Grounding; for the other tasks, the two modes perform similarly. Overall, these results align with our expectations and further guide the learning of adaptive mode selection across tasks.

\paragraph{Rollout Length in the RL Stage} Figure~\ref{fig:rl-length} illustrates the change in rollout length for two modes during the RL process. It can be observed that the lengths of rollouts in two modes are relatively close, with a difference of less than 20 tokens.

\subsection{Discussion on Scalability and Future Work}
\label{appendix:future_direction}

\begin{figure}[t]
    \centering
    \includegraphics[width=0.92\linewidth]{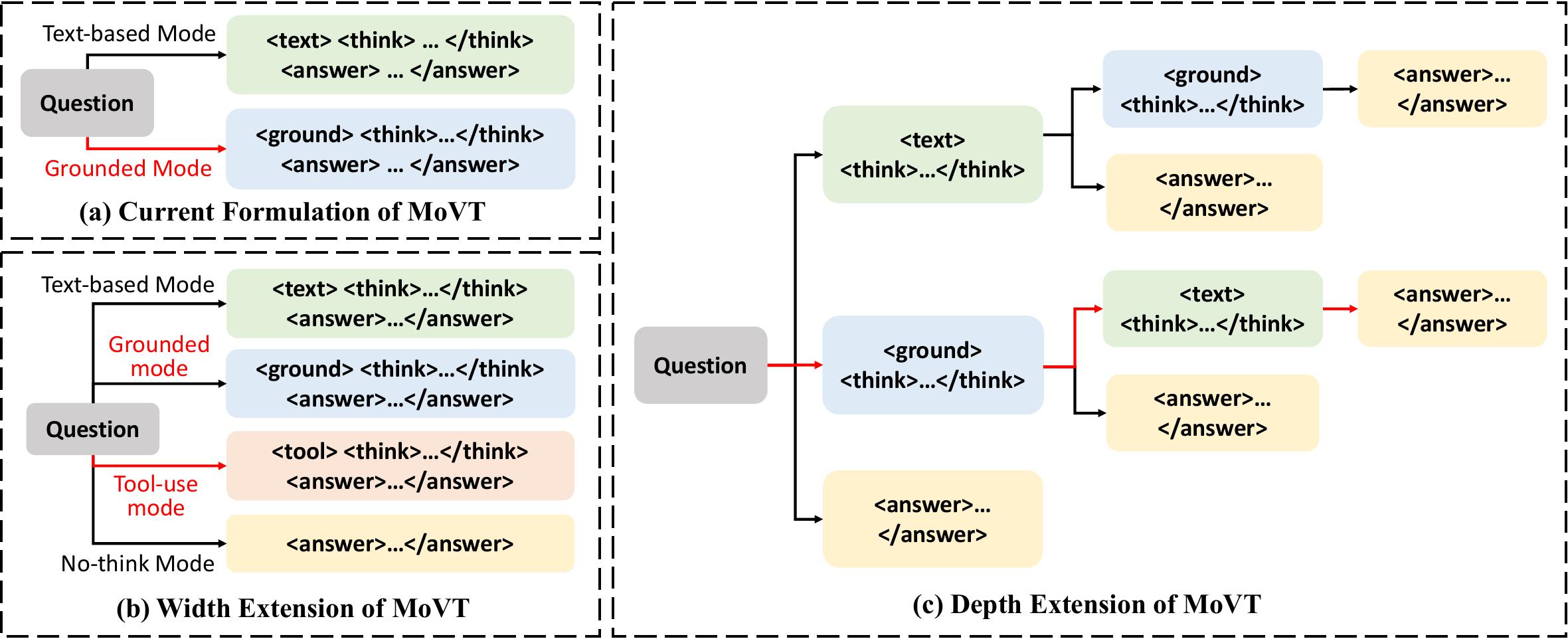}
    \caption{Extensions of MoVT to additional reasoning modes.}
    \label{fig:mvot_extension}
\end{figure}

Although this paper demonstrates that the adaptive reasoning paradigm MoVT is a feasible and effective solution for building general-purpose visual reasoning capabilities, our results further underscore its substantial potential. In this section, we discuss directions for future extension.

\subsubsection{The Scalability of MoVT To More Reasoning Modes}

In the main paper, we instantiate MVoT with two mainstream reasoning modes: a \textit{text-based} mode and a \textit{grounded} mode. This design already validates the effectiveness of MoVT as a unified paradigm for adaptive mode selection. Then, to naturally extend MVoT to support richer reasoning patterns, we consider two complementary directions illustrated in Figure~\ref{fig:mvot_extension}: \textbf{width extension} by adding more reasoning modes and \textbf{depth extension} by composing modes over multiple steps.

\paragraph{Width Extension} Beyond the current text-based and grounded modes in Figure~\ref{fig:mvot_extension}(a), MoVT can incorporate additional complementary modes by assigning each mode a dedicated prefix. For example, as shown in Figure~\ref{fig:mvot_extension}(b), we may introduce a \textbf{tool-use mode}, where the model is encouraged to invoke external image-related tools before reasoning, e.g., \texttt{<tool> <think> \ldots </think> <answer> \ldots </answer>}; or a \textbf{no-think mode}, where the model directly produces an answer without explicit intermediate reasoning, e.g., \texttt{<answer> \ldots </answer>}. 
By constructing corresponding SFT data and performing RL to compare the new modes against existing ones under the MoVT framework, new reasoning skills can be seamlessly integrated into the model while preserving the unified mode-selection mechanism.

\paragraph{Depth Extension} MoVT can also be extended in depth by allowing multi-step mode selection and reasoning, as depicted in Figure~\ref{fig:mvot_extension}(c). Concretely, at each step the model can either select a mode (e.g., text-based or grounded), performs reasoning within that mode or optionally decides to terminate by producing an answer based on the current information. This yields a flexible controller that can realize traditional single-mode reasoning as a special case but with more complex reasoning patterns constructed via sequential mixing and mode switching (e.g., grounded $\rightarrow$ text-based, or text-based $\rightarrow$ grounded). Under this formulation, inference and training can be further enhanced by search strategies and RL objectives that explicitly explore the space of multi-step mode sequences. 

% \paragraph{Extension to More Reasoning Modes} From a more general perspective, any reasoning prediction distribution induced by an inductive bias can be viewed as a specific reasoning mode—including the textual and grounded reasoning considered in this paper, long- and short-chain reasoning with respect to reasoning length, and even treating direct answering as a mode. Our AdaVaR framework is naturally extensible to additional reasoning modes: we assign a dedicated prefix token to each mode and leverage AdaGRPO to explore and learn across modes. Meanwhile, our cross-mode advantage estimation can incorporate additional preferences; for example, when performance is on par, we can prefer direct answering to improve reasoning efficiency.

% \paragraph{Inducing Complex Reasoning Modes} Introducing diverse reasoning modes expands the model’s exploratory capacity; moving forward, we can integrate search-based strategies to construct more sophisticated reasoning modes. For example, by incorporating self-consistency~\citep{wang2023selfconsistency}, we can aggregate reasoning outputs across modes; compared with relying on a single mode, integrating multiple modes enables broader exploration of the solution space. We can also enable non-linear reasoning, such as first performing grounded reasoning to analyze image content, followed by textual reasoning to produce a summary and the final answer.

\subsubsection{The Scalability of the AdaVaR Learning Framework}

The AdaVaR learning framework can also be naturally adapted to the two extension directions introduced in Appendix. While the SFT extension only requires constructing the corresponding data, here we primarily discuss the extension strategy for the AdaGRPO algorithm.

\paragraph{Scalability of Prefix-Guided Mode Exploration} For width extension to k\textgreater 2 modes, we can improve efficiency by selecting j\textless k modes to explore for each sample: (i) randomly selecting modes in the early training stages, and (ii) once the model develops certain mode selection capability, selecting the top-j modes based on the model's predicted probabilities.
For depth extension, we can intervene after the end of each thought process (i.e., after the \textless/think\textgreater~token) by append all feasible prefixes to construct rollouts with hierarchical and thorough exploration.

\paragraph{Advancing Mode Selection} As shown in Table~\ref{table:adaptive_analysis} and Appendix~\ref{appendix:error_analysis}, although AdaVaR achieves substantial performance gains, there remains a gap to the upper bound of adaptive reasoning, indicating that the model’s mode-selection capability can be further improved. In the future work, we can explore to construct richer datasets, or introduce a dedicated reasoning mechanism for mode selection itself—having the model first reason about which mode to use and then reason about the problem—to strengthen mode selection.

\subsubsection{Extension to Wider Domains}

\paragraph{Extension to Multilingual Context} As stated in Ethics Statement, following mainstream visual reasoning models, our work primarily focuses on English scenarios. The main limitation preventing thorough exploration of multilingual visual reasoning is the lack of high-quality data, both for training and evaluation. In future work, we aim to explore several key directions: (1) how to efficiently construct multilingual reasoning-oriented datasets; (2) the cross-lingual transferability and generalization of reasoning capabilities; and (3) how to develop adaptive reasoning methods for multilingual scenario, in order to extend MoVT and AdaVaR to a broader range of languages.

\paragraph{Extension to Other Modalities} Our method primarily focuses on images. The improvements observed on video and multi-view 3D scenes in SpatialScore indicate that our approach can be extended to additional modalities. When dealing with diverse modalities—such as video, 3D data, and medical imaging—we believe adaptive reasoning will offer even greater value, as different modalities often require distinct reasoning modes. In future work, we aim to explore grounded reasoning methods in high-dimensional scenarios, expand the repertoire of reasoning modes, and investigate the transferability of these modes across different modalities.

\subsection{Supplementary Ablation Study}
\label{appendix:ablation}

\input{tables/sft_data_ratio}

\input{tables/mode_prefix_ablation}

\subsubsection{The 1:1 Data Ratio between Two Modes during SFT}

As stated in Section~\ref{section:SFT_stage}, we chose the 1:1 ratio to avoid introducing a preference for any specific mode during the SFT stage. To validate this design, in Table~\ref{table:ablation_sft_data_ratio}, we experiment with different data ratios under the premise of keeping the total amount of data constant. 

Several findings can be gleaned: (1) Higher proportion of a specific mode in the SFT data makes the model more inclined to select that mode: this tendency is particularly evident in the SFT model. After RL, different models converge toward similar trends, but the influence of the SFT stage still persists.
(2) Such a preference may be effective on specific datasets (e.g., the preference for the text mode improves performance on WeMath); however, from the perspective of generalization (average performance), although all biased settings outperform the baseline, none perform as well as the 1:1 setting. (3) In addition, a major limitation of applying a skewed data ratio is that, in practical applications, facing various queries, it is difficult to know in advance which mode would be better.
In summary, although the RL phase helps reduce the sensitivity to the data in the SFT stage, an unbiased design between different modes is optimal for achieving superior general performance.

\subsubsection{The Format of Mode Prefixes}

Currently, we utilize two mode prefixes with explicit semantics. To investigate the sensitivity to the format of prefixes, we conduct an ablation experiment, using semantically neutral prefixes, namely \textless mode1\textgreater~and \textless mode2\textgreater. As presented in Table~\ref{table:ablation_on_mode_prefix}, models with different prefix formats show similar performance, indicating that, within our mode unification design, different prefix formats have minimal impact, and the model is able to learn distinct reasoning modes after SFT.

\input{tables/sft_ablation}
\input{tables/system_prompt}

\subsubsection{The Effects of System Prompt Design}
In our framework, we learn different reasoning modes during the SFT stage, where supervision primarily come from the SFT data, and the system prompt serves as an auxiliary guide.
To verify the role of the system prompt, we conduct two ablation studies, aiming to address two questions: (1) Can different reasoning modes be elicited solely through the system prompt without relying on SFT data? and (2) Does the system prompt require detailed descriptions of the modes?

For the first question, we ablate AdaVaR-3B by skipping the SFT stage. Results in Table~\ref{table:ablation_sft_stage} show that:
(i) Even provided with the description of the grounded mode, the model does not elicit the ability to perform grounded reasoning during the RL phase—only 0.03\% of rollouts contain coordinates.
(ii) In evaluations, the model shows no clear distinction between modes and fails to generate responses in the grounded reasoning format, achieving only a marginal improvement over Qwen2.5-VL-3B.
Therefore, these results show that it is difficult to directly elicit multiple distinct reasoning modes merely by designing system prompts, which highlights the necessity of the SFT phase in our framework—regardless of whether it is applied to Qwen2.5-VL or other LVLMs.

To investigate the sensitivity to the system prompt, we experiment with two variants: \textit{full}, which includes both the thinking instructions and the descriptions of the modes, and \textit{think-only}, which removes the mode descriptions. Table~\ref{table:system_prompt} presents the results obtained using different system prompts during training and evaluation: (i) Our model, that is trained with the full prompt, can directly transfer to the think-only prompt and still maintain the capabilities of both reasoning modes, performing reasonable mode selection.
(ii) When training with the think-only prompt, the model achieves performance nearly on par with that of the model trained using the full prompt.
These results indicate that, after SFT, the model exhibits low sensitivity to the prompt. When extending to more modes, there is no need to design complex prompts, which enhances the scalability of our framework.

\input{tables/data_order}

\subsubsection{Data Order For Curriculum Learning}

Results in Table~\ref{table:ablation_study} validate the effectiveness of the easy-to-hard data order. In this section, we investigate whether a hard-to-easy order works. As illustrated in Table~\ref{table:ablation_on_data_order}, the model trained with a hard-to-easy data order consistently underperforms compared to the easy-to-hard setting, validating the rationality of our design.

\input{tables/gaussian_blur}

\subsubsection{Robustness against Image Noise}

To investigate whether the adaptive mode selection ability in AdaVaR is sensitive to image noises, we conduct an experiment by applying Gaussian blur to images during evaluation, we set the radius of blur to 0.5, in order to prevent excessive degradation of image information. The results are shown in Table~\ref{table:image_blur}, we find that: (1) introducing noise does not significantly affect mode selection; (2) the overall performance of AdaVaR shows only a slight decline; (3) even with added noise, AdaVaR-3B still maintains a clear advantage over baselines. These supports the robustness of our model.

\subsection{Qualitative Analysis}
\label{appendix:case_study}

To illustrate the characteristics of different reasoning modes and how our AdaVaR model select the appropriate mode, we provide examples across multiple datasets, including image-related questions, the reasoning trajectories of different modes, and the model’s mode selections.

\paragraph{General Notation and Findings} In this section, all figures follow the same notation scheme: we underline the \underline{correct answers} and visualize in the images the bounding boxes generated by the visually-grounded mode; different colors indicate the correspondence between the bounding boxes in the images and in the text; ``TXT Mode'' and ``GRD Mode'' are respectively short for the text-based mode and the visually-grounded mode. By examining cases across different datasets, we observe clear differences in the reasoning trajectories produced by the two modes, which also validates our discussion in Section~\ref{section:delving_question_1}: the two modes induce distinct reasoning patterns. In general, text-based reasoning is better at constructing complex, abstract logic, whereas the grounded mode tends to locate and search for key information and analyze it.

\begin{figure}[t]
    \centering
    \includegraphics[width=0.96\linewidth]{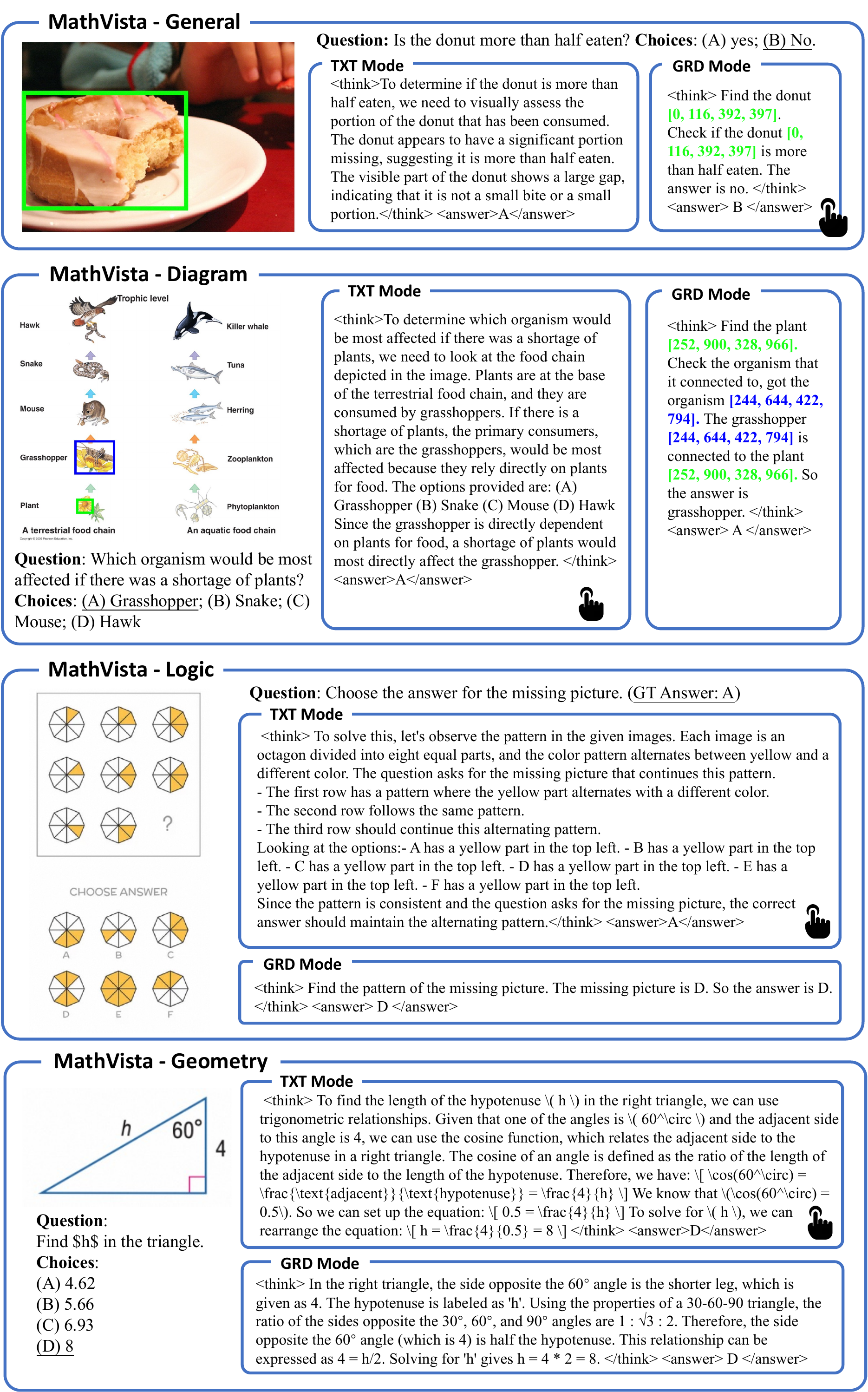}
    \caption{Cases from MathVista. The pointing-finger icon indicates the mode selected by AdaVaR.}
    \label{fig:mathvista_cases}
\end{figure}

\paragraph{MathVista} MathVista includes both math-oriented and general-scenario questions. As presented in Figure~\ref{fig:mathvista_cases}, we make several observations: (1) For more general scenarios—for example, the first case—the grounded mode, which places greater emphasis on visual information, performs better; (2) On tasks with relatively simple solution methods, such as the second case, the grounded mode also succeeds; (3) For problems requiring complex and abstract logic, like the third case, the textual mode is superior; (4) Moreover, when the grounded mode recognizes that there are no concepts requiring localization—for instance, the geometry problem in the fourth case—it also exhibits a certain degree of abstract reasoning ability.

\begin{figure}[t]
    \centering
    \includegraphics[width=0.96\linewidth]{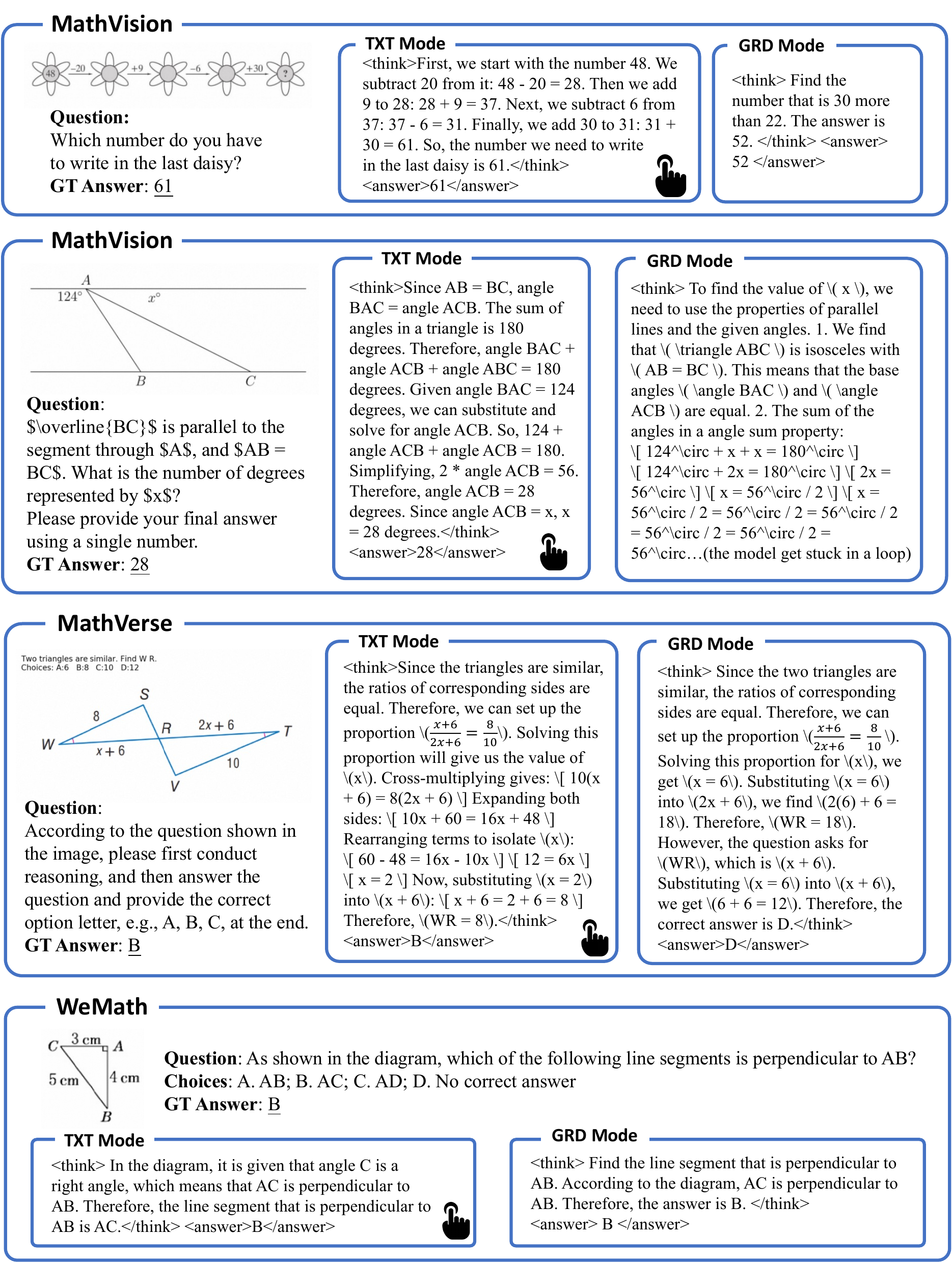}
    \caption{Cases from other math-oriented benchmarks.}
    \label{fig:math_cases}
\end{figure}

\paragraph{Math-Oriented Benchmarks} Figure~\ref{fig:math_cases} shows examples from three other math-oriented datasets. We observe a pattern: in these datasets, for nouns referring to abstract concepts (e.g., numbers), the grounded reasoning mode may invoke the “find” mechanism yet still cannot locate a concrete corresponding bounding box in the image. This corroborates our assumption that performing visual grounding is difficult in abstract scenarios. In such cases, the grounded mode can handle simple questions (e.g., the fourth case), but fails on those requiring step-by-step computation, such as the first case. Another observation is that the grounded mode is weak at numerical computation. For example, in the second and third cases it produces correct initial derivations but makes mistakes during the numerical calculations; in the second case, it even gets stuck in a repetitive generation loop and fails to solve the corresponding equation.

\begin{figure}[t]
    \centering
    \includegraphics[width=0.93\linewidth]{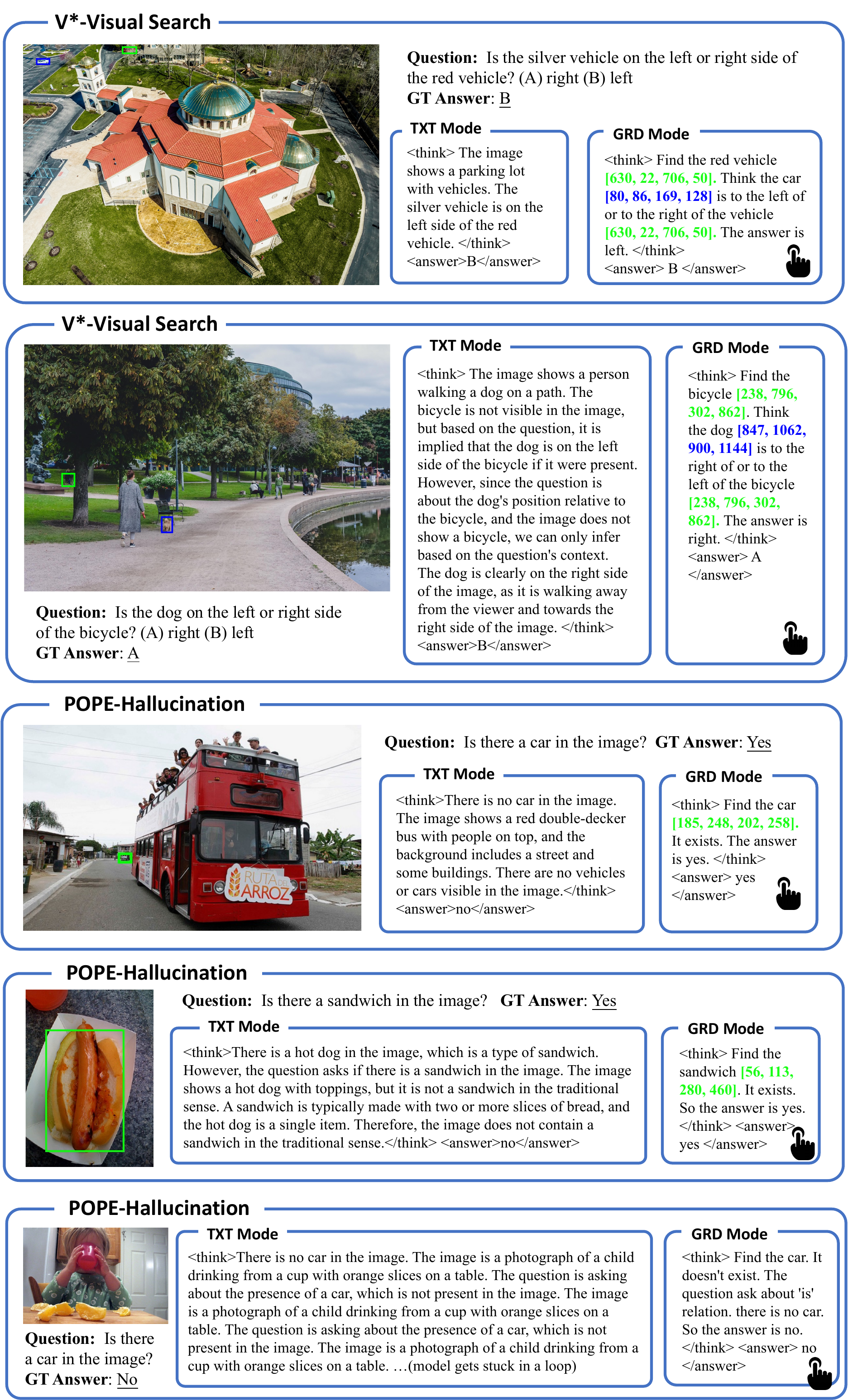}
    \caption{Cases from V* and POPE. Please zoom in for a better view of small objects in images.}
    \label{fig:vstar_pope_cases}
\end{figure}

\paragraph{V*} V* and POPE are representative of tasks that hinge on fine-grained object-related details. V* in particular tests the model’s ability to localize small objects. We provide several cases in Figure~\ref{fig:vstar_pope_cases}. As seen in the first two cases, text-based reasoning in such scenarios yields reasoning that lacks interpretability: without coordinates, users cannot tell whether the model actually localized the target object. Text-based reasoning is also prone to hallucinations; for example, in the second case the model produces incorrect information: it fails to find the bicycle and assumes in advance that the dog is to the left of the bicycle. In contrast, grounded reasoning can precisely locate the target object in such complex images, and the correct coordinates help the model answer the question correctly.

\paragraph{POPE} Considering the hallucination task in POPE, the third case in Figure~\ref{fig:vstar_pope_cases} shows that text-based reasoning mode struggles to locate the target object (especially when it is small, this limitation is also evident in V*). Moreover, text-based reasoning can lead to overthinking—for example, the misunderstanding of a sandwich in the fourth case, and the introduction of irrelevant information and a repetitive generation loop in the fifth case. In contrast, the grounded mode thoroughly checks the objects in the image and provides correct answers.

\begin{figure}[t]
    \centering
    \includegraphics[width=0.95\linewidth]{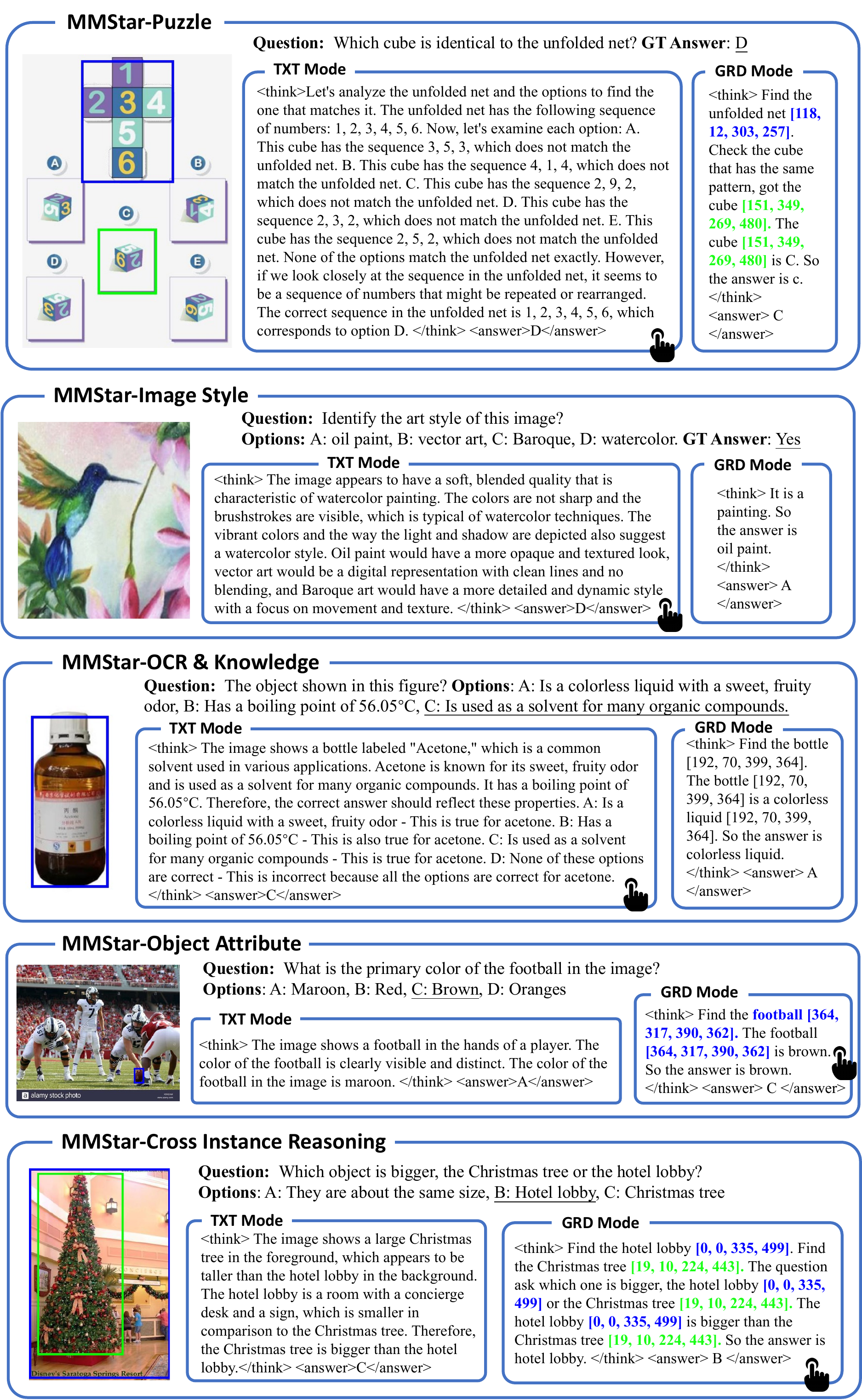}
    \caption{Cases from MMStar.}
    \label{fig:mmstar_cases}
\end{figure}

\paragraph{MMStar} As a benchmark including samples across different scenarios, We find that in many scenarios, both reasoning modes can arrive at the correct answer, so in Figure~\ref{fig:mmstar_cases} we present several examples where the two modes differ. According to the first three examples, we can see that the text-based reasoning mode is better suited to tasks requiring complex abstract logic and knowledge—for instance, the folding problem in the first example and the painting-style knowledge in the second. The third example shows that the grounded mode tends to leverage information visible in the image, such as the colorless liquid, but cannot further infer additional information based on ``acetone''. The last two cases show that the grounded mode is better at capturing on attributes of specific objects and making comparisons between target objects.

\begin{figure}[t]
    \centering
    \includegraphics[width=0.92\linewidth]{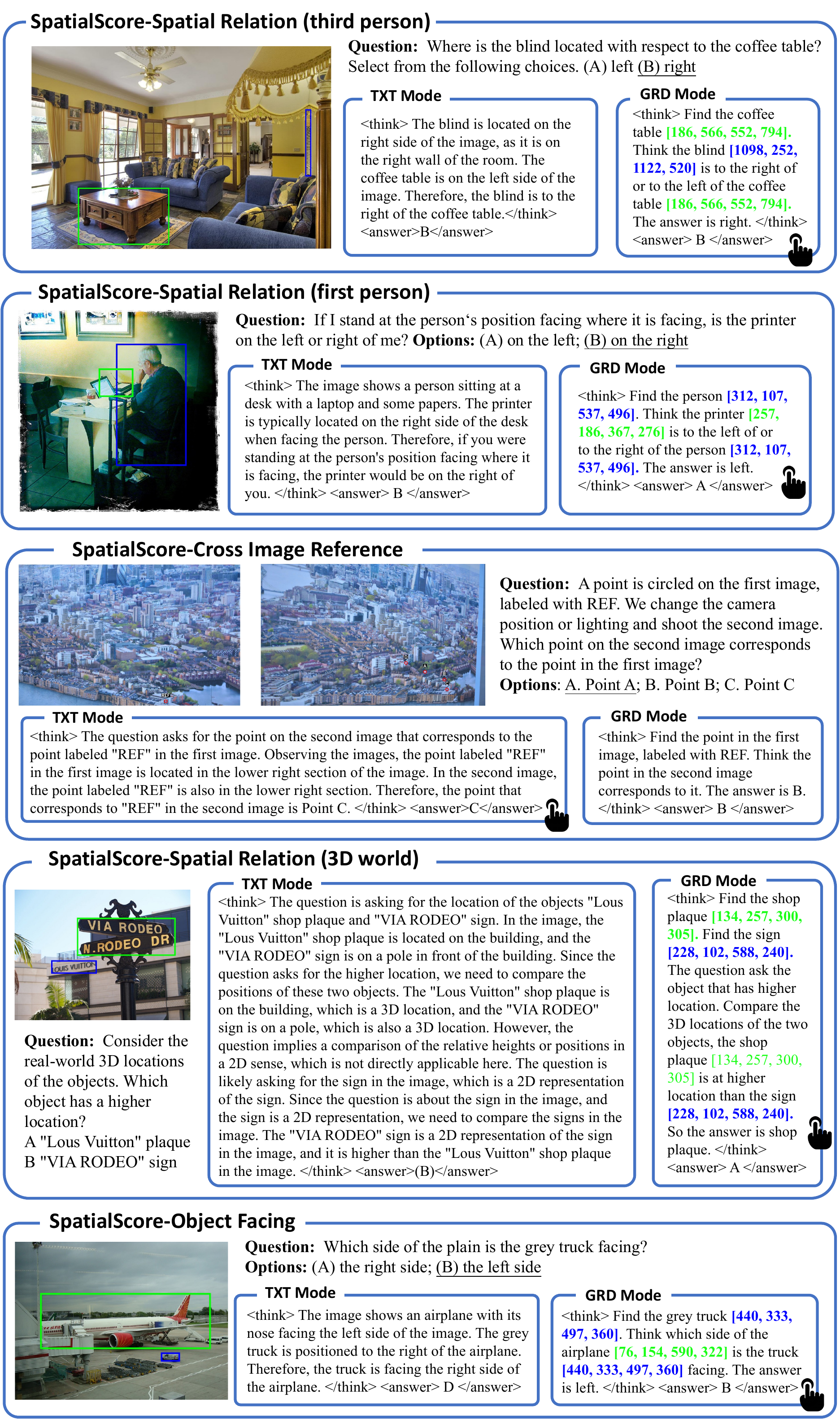}
    \caption{Cases from SpatialScore.}
    \label{fig:spatialscore_cases}
\end{figure}

\paragraph{SpatialScore} Focusing on spatial reasoning in general scenes, Figure~\ref{fig:spatialscore_cases} provides several examples. The first case is a relatively common one from a third-person perspective (camera viewpoint). Both reasoning modes answer correctly, and the grounded mode further indicates the specific locations of the objects. The second case further requires reasoning about spatial relations from the first-person perspective of a specific object, which calls for some imagination. Thus, although the grounded mode correctly locates the corresponding object, it lacks further imaginative reasoning and does not answer correctly as the textual mode does. In addition, the third case reveals a current limitation of the grounded mode in multi-image scenarios: it cannot distinguish between different images using coordinates, which is a direction for future improvement. The fourth and fifth cases illustrate the grounded mode’s advantage on more conventional spatial reasoning problems: by locating key objects and combining common sense, both ultimately yield correct reasoning outcomes.

\begin{figure}[t]
    \centering
    \includegraphics[width=0.92\linewidth]{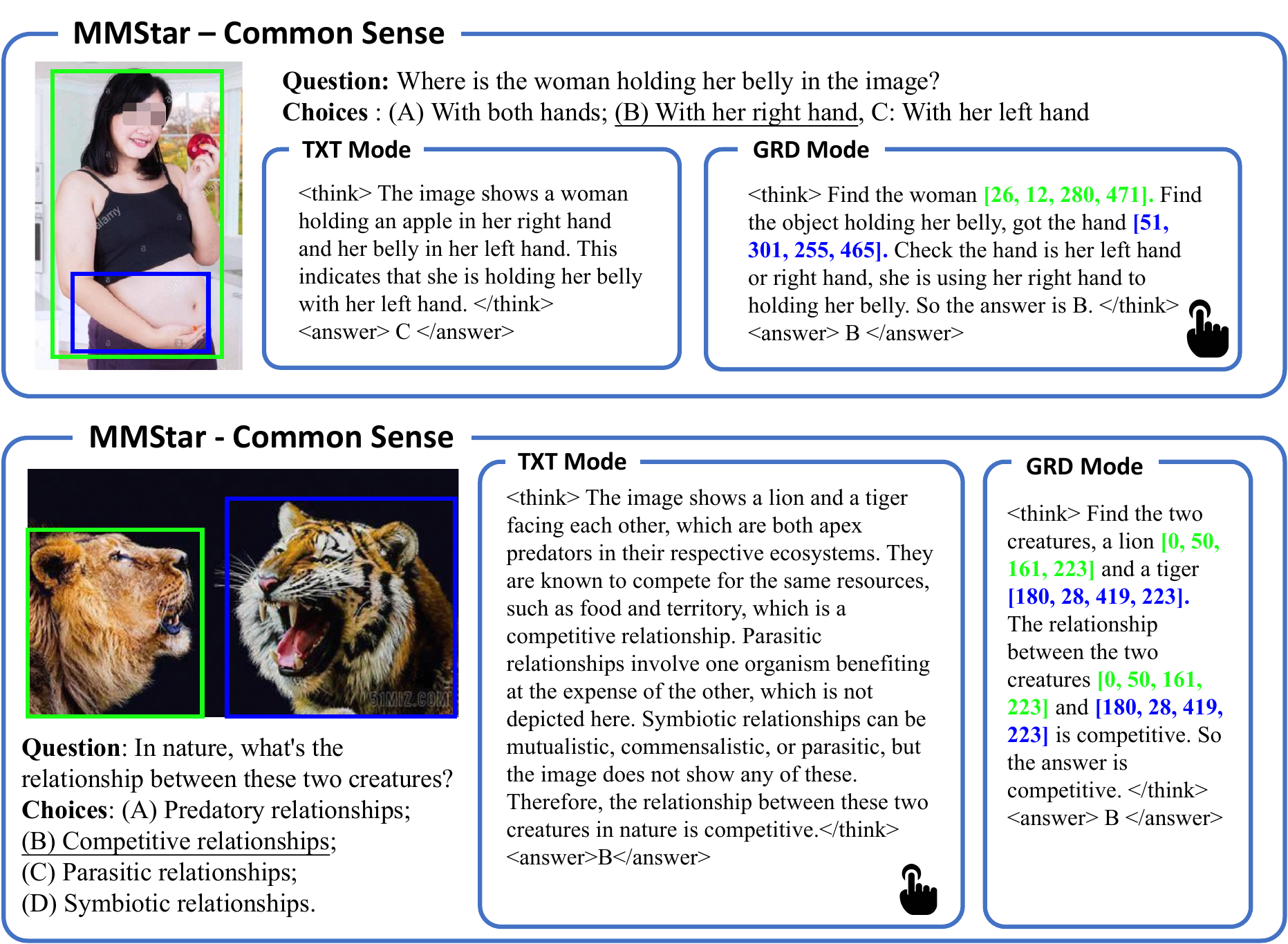}
    \caption{Cases from the common sense reasoning task in MMStar.}
    \label{fig:mmstar_rebuttal_cases}
\end{figure}

\paragraph{Extra Qualitative Results on MMStar}
We provide two additional qualitative examples from the common-sense reasoning subset of MMStar in Figure~\ref{fig:mmstar_rebuttal_cases}. These cases require diverse focus: not only grounding on relevant objects in the image but also performing abstract common-sense reasoning. Rather than relying on a single mode, our model dynamically selects the most suitable mode for each question.

% \section{Rebuttal Materials}

% In this section, we include additional figure-related analysis we introduced during the rebuttal phase.

% \subsection{Discussion on the Scalability}

% \subsubsection{The extension of MVoT to more reasoning modes}

\end{document}

%% file: math_commands.tex
\usepackage{amsmath,amsfonts,bm}

\def\eqref#1{equation~\ref{#1}}
\def\1{\bm{1}}

\DeclareMathAlphabet{\mathsfit}{\encodingdefault}{\sfdefault}{m}{sl}
\SetMathAlphabet{\mathsfit}{bold}{\encodingdefault}{\sfdefault}{bx}{n}

%% file: tables/main_table.tex
% Please add the following required packages to your document preamble:
% \usepackage{multirow}
\begin{table}[t]
\caption{Evaluation results on various benchmarks. * indicates results reproduced in this paper. Within each model group with different parameter sizes, the best and second-best results are highlighted in \textbf{bold} and \underline{underlined}, respectively. We use background colors to distinguish \colorbox{layer1}{text-based}, \colorbox{layer2}{visually-grounded}, and \colorbox{layer3}{our adaptive} reasoning models. ``Acc.'' is short for accuracy.}
\vspace{-2mm}
\label{table:main_results}
\resizebox{1\textwidth}{!}{
\begin{tabular}{lccccccccc}
\toprule
\multicolumn{1}{l|}{\multirow{2}{*}{\textbf{Model}}} & \multicolumn{4}{c|}{\textbf{Math-Oriented}}                                                                                                                                                                                & \multicolumn{4}{c|}{\textbf{General Scenarios}}                                                                                                                   & \multirow{3}{*}{\textbf{\begin{tabular}[c]{@{}c@{}}Average\\ Acc.\end{tabular}}} \\ \cmidrule{2-9}
\multicolumn{1}{l|}{}                                & \begin{tabular}[c]{@{}c@{}}\textbf{MathVista}\\ testmini\end{tabular} & \textbf{MathVision} & \begin{tabular}[c]{@{}c@{}}\textbf{MathVerse}\\ vision-only\end{tabular} & \multicolumn{1}{c|}{\begin{tabular}[c]{@{}c@{}}\textbf{WeMath}\\ strict\end{tabular}} & \textbf{MMStar} & \textbf{V*}     & \begin{tabular}[c]{@{}c@{}}\textbf{POPE}\\ all\end{tabular} & \multicolumn{1}{c|}{\begin{tabular}[c]{@{}c@{}}\textbf{SpatialScore}\\ hard\end{tabular}} &                                                                                  \\ \midrule
\rowcolor[HTML]{EFEFEF}
\multicolumn{10}{c}{SFT \& Close-source Models}                                                                                                                                                                                                                                                                                                                                                                                                                                                                                \\ \midrule
\multicolumn{1}{l|}{GPT-4o~\citeyearpar{hurst2024gpt}}                          & 63.8                                                         & 30.4       & 39.9                                                            & \multicolumn{1}{c|}{42.9}                                                    & 64.7   & 66.0     & 86.9                                               & \multicolumn{1}{c|}{30.6}                                                       & 53.20                                                                         \\
% \multicolumn{1}{l|}{GPT-4o mini~\citeyearpar{hurst2024gpt}}                     & 55.1                                                         & 27.3       &  -                                                               & \multicolumn{1}{c|}{31.4}                                                    & -       & -       & -                                                   & \multicolumn{1}{c|}{-}                                                            &  -                                                                                \\
\multicolumn{1}{l|}{LLaVA-OV-7B~\citeyearpar{li2024llavaonevisioneasyvisualtask}}              & 62.6                                                         & 17.6       & 19.3                                                            & \multicolumn{1}{c|}{17.7}                                                    &  -      & 75.4   & 88.4                                               & \multicolumn{1}{c|}{15.6}                                                        &  -                                                                                \\
% \multicolumn{1}{l|}{Qwen2-VL-7B~\citeyearpar{wang2024qwen2vlenhancingvisionlanguagemodels}}                     & 61.6                                                         & 19.2       &  -                                                               & \multicolumn{1}{c|}{22.3}                                                    & -       & -      & 88.1                                               & \multicolumn{1}{c|}{-}                                                           & -                                                                                 \\
\multicolumn{1}{l|}{InternVL3-8B~\citeyearpar{zhu2025internvl3}}                    & 70.5                                                         & 28.6       &  39.8                                                               & \multicolumn{1}{c|}{37.5}                                                    & 66.3   & -       & 91.1                                               & \multicolumn{1}{c|}{12.9}                                                       &  -                                                                                \\ \midrule
\rowcolor[HTML]{EFEFEF}
\multicolumn{10}{c}{Reasoning Models based on Qwen2.5-VL-3B}                                                                                                                                                                                                                                                                                                                                                                                                                                                                   \\ \midrule
\multicolumn{1}{l|}{Qwen2.5-VL-3B~\citeyearpar{bai2025qwen25vltechnicalreport}}                   & 62.3                                                         & 21.2       & 31.2*                                                            & \multicolumn{1}{c|}{21.2*}                                                   & 53.1* & 73.5* & 87.2*                                             & \multicolumn{1}{c|}{16.2}                                                       & 45.74                                                                            \\
\rowcolor[HTML]{FEFEDA}
\multicolumn{1}{l|}{LMM-R1~\citeyearpar{peng2025lmm}}                          & \underline{63.2}                                                         & \textbf{25.2}       & \textbf{37.3}*                                                            & \multicolumn{1}{c|}{\underline{29.2*}}                                                   & \underline{55.0*}     & 74.9*   & 86.7*                                               & \multicolumn{1}{c|}{18.9*}                                                        & \underline{48.80}                                                                           \\
\rowcolor[HTML]{FEFEDA}
\multicolumn{1}{l|}{VLAA-Thinker-3B~\citeyearpar{chen2025sft}}                 & 61.0                                                           & 24.4       & \underline{36.4}                                                            & \multicolumn{1}{c|}{\textbf{33.8}}                                                    & 54.8*   & 56.0*  & 87.8*                                               & \multicolumn{1}{c|}{\underline{19.4*}}                                                        & 46.70      \\                                                                    
\rowcolor[HTML]{EDFEED} 
\multicolumn{1}{l|}{GRIT-3B~\citeyearpar{fan2025gritteachingmllmsthink}}                         & 59.8                                                         & 21.2*       & 28.4*                                                            & \multicolumn{1}{c|}{29.1*}                                                    & 52.7*   & 72.3*   & 83.0*                                                 & \multicolumn{1}{c|}{\textbf{19.7}*}                                                        & 45.80                                                                           \\
\rowcolor[HTML]{EDFEED} 
\multicolumn{1}{l|}{ViGoRL-3B~\citeyearpar{sarch2025grounded}}                       &   55.2*                                                           & 18.9*           &  24.2*                                                               & \multicolumn{1}{c|}{28.5*}                                                        & 51.4*       & \textbf{78.1}   &  \underline{88.0*}                                                  & \multicolumn{1}{c|}{18.2*}                                                            & 45.70                                                                             \\
\rowcolor[HTML]{EFF6FE}
\multicolumn{1}{l|}{AdaVaR-3B (Ours)}                       & \textbf{69.8}                                                         & \underline{24.5}       & 35.2                                                            & \multicolumn{1}{c|}{\textbf{33.8}}                                                   & \textbf{59.3}   & \underline{77.0}     & \textbf{88.2}                                               & \multicolumn{1}{c|}{18.9}                                                        & \textbf{50.84}                                                                           \\ \midrule
\rowcolor[HTML]{EFEFEF}
\multicolumn{10}{c}{Reasoning Models based on Qwen2.5-VL-7B}                                                                                                                                                                                                                                                                                                                                                                                                                                                                   \\ \midrule
\multicolumn{1}{l|}{Qwen2.5-VL-7B~\citeyearpar{bai2025qwen25vltechnicalreport}}                   & 68.2                                                         & 25.1       & 41.1                                                            & \multicolumn{1}{c|}{31.2}                                                    & 60.3* & 78.0* & 87.8*                                              & \multicolumn{1}{c|}{15.2}                                                       & 50.90                                                                             \\
\rowcolor[HTML]{FEFEDA}
\multicolumn{1}{l|}{VLAA-Thinker-7B~\citeyearpar{chen2025sft}}                 & 68.0                                                           & 26.4       & \underline{48.2}                                                            & \multicolumn{1}{c|}{41.5}                                                    & 62.6*   & 78.5*  & 86.2*                                               & \multicolumn{1}{c|}{22.3*}                                                       & 54.20                                                                          \\
\rowcolor[HTML]{FEFEDA}
\multicolumn{1}{l|}{MM-Eureka~\citeyearpar{meng2025mmeurekaexploringfrontiersmultimodal}}                       & \underline{72.6}                                                         & 28.1       & 45.4                                                            & \multicolumn{1}{c|}{36.9}                                                    & \textbf{64.0}*     & 59.7*   & 86.3*                                               & \multicolumn{1}{c|}{\textbf{27.1}*}                                                        & 52.52                                                                           \\
\rowcolor[HTML]{FEFEDA}
\multicolumn{1}{l|}{OVR-7B~\citeyearpar{wei2025open}}                          & 72.1                                                         & \textbf{46.4}*       & \textbf{54.6}                                                            & \multicolumn{1}{c|}{\underline{44.6}}                                                    & 62.7   & 62.5*   & 83.2                                               & \multicolumn{1}{c|}{20.0*}                                                          & \underline{55.76}                                                                          \\
\rowcolor[HTML]{FEFEDA}
\multicolumn{1}{l|}{Orsta-7B~\citeyearpar{ma2025one}}                        & 72.5                                                         & 28.2*       & 42.9*                                                            & \multicolumn{1}{c|}{31.8*}                                                    & 59.6*   & 78.0*     & 86.9*                                               & \multicolumn{1}{c|}{16.4*}                                                        & 52.04                                                                          \\
\rowcolor[HTML]{EDFEED} 
\multicolumn{1}{l|}{DeepEyes~\citeyearpar{zheng2025deepeyesincentivizingthinkingimages}}                        & 70.1                                                         & 26.6       & 40.7*                                                         & \multicolumn{1}{c|}{32.7*}                                                    & 61.3*       & \textbf{90.1}  & 87.9*                                               & \multicolumn{1}{c|}{20.3*}                                                            &  53.72                                                                                \\
\rowcolor[HTML]{EDFEED} 
\multicolumn{1}{l|}{Chain-of-Focus~\citeyearpar{zhang2025chainoffocusadaptivevisualsearch}}                  & 63.1*                                                             & 22.9*           & 32.1*                                                                & \multicolumn{1}{c|}{33.1*}                                                        & 59.3*       & \underline{88.0}     & \underline{88.4}                                               & \multicolumn{1}{c|}{\underline{20.6*}}                                                            &  50.94                                                                                \\
\rowcolor[HTML]{EDFEED} 
\multicolumn{1}{l|}{ViGoRL-7B~\citeyearpar{sarch2025grounded}}                       & 63.5*                                                             & 23.3*           &  32.6*                                                               & \multicolumn{1}{c|}{36.3*}                                                        & 54.3*       & 86.4       &  88.3*                                                  & \multicolumn{1}{c|}{19.5*}                                                            & 50.53                                                                                 \\
\rowcolor[HTML]{EFF6FE}
\multicolumn{1}{l|}{AdaVaR-7B (Ours)}                       & \textbf{74.4}                                                         & \underline{28.5}       & 43.0                                                           & \multicolumn{1}{c|}{\textbf{44.8}}                                                    & \underline{63.0}     & 83.4   & \textbf{89.0}                                                 & \multicolumn{1}{c|}{20.4}                                                       & \textbf{55.82}                                                                         \\ \bottomrule
\end{tabular}}
\vspace{-6mm}
\end{table}

%% file: tables/delve_adaptive_v2.tex
\begin{table}[t]
\caption{Performance of AdaVaR with different reasoning modes at various stages. AdaVaR without a subscript performs adaptive reasoning. AdaVaR with the subscripts $G$ and $T$ denote the performance using visually-grounded mode and text-based mode, respectively. GRD\% indicates the proportion of times AdaVaR chooses the grounded mode in the corresponding dataset.}
\label{table:adaptive_analysis}
\vspace{-2mm}
\resizebox{0.98\textwidth}{!}{
\begin{tabular}{l|cccc|cccc|c}
\toprule
\multirow{2}{*}{\textbf{Model}} & \multicolumn{4}{c}{\textbf{Math-Oriented}}  & \multicolumn{4}{|c|}{\textbf{General Scenarios}}          & \multirow{2}{*}{\textbf{\begin{tabular}[c]{@{}c@{}}Average\\ Acc.\end{tabular}}} \\ \cmidrule{2-9}
                                & MathVista & MathVision & MathVerse & WeMath & MMStar & V*   & POPE  & SpaScore &                                                                                  \\ \midrule
Qwen2.5-VL-3B                   & 62.3      & 21.2       & 31.2      & 21.1   & 53.1      & 73.5 & 86.1  & 16.2         & 45.6                                                                             \\ \midrule
\multicolumn{10}{c}{Stage 1 SFT-based Models} \\ \midrule
AdaVaR-SFT-3B                   & 60.4      & 23.2       & 33.0      & 29.5   & 54.7     & 75.5 & 89.1  & 16.8         & 47.8                                                                             \\
AdaVaR-SFT-3B$_{G}$                & 56.4      & 20.3       & 28.5      & 29.5   & 53.2     & 75.5 & 89.2  & 17.6         & 46.3                                                                             \\
AdaVaR-SFT-3B$_T$                & 62.5      & 22.9       & 31.4      & 31.1   & 55.9    & 65.6 & 83.1  & 17.8         & 46.3                                                                             \\
GRD\%                           & 31\%      & 5\%        & 1\%       & 78\%   & 90\%    & 98\% & 94\%  & 96\%         & -                                                                                \\ \midrule
\multicolumn{10}{c}{Stage 2 RL-based Models} \\ \midrule

AdaVaR-3B                       & 69.8      & 24.5       & 35.2      & 33.8   & 59.3    & 77.1   & 88.2  & 18.9         & 50.8                                                                             \\
AdaVaR-3B$_G$                    & 66.1      & 22.1       & 33.9      & 32.1       & 56.3     & 76.4 & 88.2  & 18.5         & 49.3                                                                                \\
AdaVaR-3B$_T$                    & 68.1      & 23.0       & 34.9      & 32.6   & 58.5    & 58.6 & 84.6  & 17.7         & 47.3                                                                                \\
Upper Bound                     & 74.8      & 30.9       & 41.7      &  38.3      & 67.6     & 78.5 & 95.6  & 27.8         & 56.8                                                                                \\
GRD\%                           & 2\%      & 0\%      & 0\%     & 0\%  & 51\%     & 99\% & 100\% & 59\%       & -                                                                                \\ \midrule
\multicolumn{10}{c}{Baseline Models based on a Single Reasoning Mode} \\ \midrule
Grounded-SFT-RL                      & 62.6      & 22.2      & 34.4      & 33.0     & 55.4         & 77.0   & 87.4  & 18.4         & 48.7  \\
Text-SFT-RL                     & 67.2      & 23.5      & 36.1      & 32.2  & 56.0           & 70.7 & 85.6  & 18.8         &  48.8   \\
Mix-SFT-RL & 65.0  & 23.2  & 33.6  & 28.8  & 56.2  & 76.9  & 85.4  & 18.6  & 48.5
\\ \bottomrule
\end{tabular}}
\vspace{-2mm}
\end{table}

%% file: tables/ablation.tex
\begin{table}[t]
\caption{Ablation study. Ada-Adv and PG-Exp are abbreviation for adaptive advantage and prefix-guided exploration, respectively. The definition of diverse mixed data is provided in Section~\ref{section:rl_data}.}
\label{table:ablation_study}
\vspace{-2mm}
\resizebox{1\textwidth}{!}{
\begin{tabular}{l|cccc|cccc|c}
\toprule
\multirow{2}{*}{\textbf{Model}} & \multicolumn{4}{c}{\textbf{Math-Oriented}}  & \multicolumn{4}{|c|}{\textbf{General Scenarios}}          & \multirow{2}{*}{\textbf{\begin{tabular}[c]{@{}c@{}}Average\\ Acc.\end{tabular}}} \\ \cmidrule{2-9}
                                & MathVista & MathVision & MathVerse & WeMath & MMStar & V*   & POPE  & SpaScore &                                                                                  \\ \midrule
AdaVaR-3B                       & 69.8      & 24.5       & 35.2      & 33.8   & 59.3    & 77.1   & 88.2  & 18.9         & 50.8                                                                             \\ \midrule
w/o Ada-Adv + PG-Exp                & 66.3      & 23.8       & 34.9      & 31.3   & 56.6     & 75.4 & 89.1  & 19.7         & 49.6                                                                          
\\
w/o Ada-Adv                   & 68.4      & 24.7       & 34.4      & 33.7   & 58.9     & 77.4 & 88.0  & 17.4         & 50.3                                                                             \\ \midrule
w/o Diverse Mixed Data                    & 67.4      & 23.7       & 33.1      & 33.4       & 57.3     & 76.4 & 82.1  & 18.5         & 49.0                                                                                \\
w/o Curriculum Learning                    & 66.8      & 24.4       & 34.9      & 33.4   & 57.8    & 76.9 & 88.2  & 18.5         & 50.1  
\\ \bottomrule
\end{tabular}}
\vspace{-2mm}
\end{table}

%% file: tables/mode_selection_image_types.tex
\begin{table}[t]
\centering
\caption{The proportion of mode selected by AdaVaR-3B across different image types in MMStar.}
\vspace{-2mm}
\begin{tabular}{l|cccccc}
\toprule
Mode Rate  & Natural Images & Art     & Document & Diagram & Geometry & Others \\ \midrule
Grounded   & 92.20\%        & 39.40\% & 8.10\%   & 4.50\%  & 0\%      & 0\%    \\
Text-based & 7.80\%         & 60.60\% & 91.90\%  & 95.50\% & 100\%    & 100\%  \\ \bottomrule
\end{tabular}
\label{table:mode_select_image_types}
\vspace{-2mm}
\end{table}

%% file: tables/naive_ensemble.tex
\begin{table}[t]
\caption{Comparison between AdaVaR and simple mode ensemble strategies.}
\vspace{-2mm}
\resizebox{0.98\textwidth}{!}{
\begin{tabular}{l|cccccccc|c}
\toprule
Model         & MathVista & MathVision & MathVerse & WeMath & MMStar & POPE & V*   & SpaScore & Avg. Acc \\ \midrule
AdaVaR-3B     & 69.8      & 24.5       & 35.2      & 33.8   & 59.3   & 77.1 & 88.2 & 18.9         & 50.8     \\
Expert Pick   & 68.1      & 23.0       & 34.9      & 32.6   & 56.3   & 76.4 & 88.2 & 18.5         & 49.8     \\
Max(GRD, TXT) & 68.1      & 23.0       & 34.9      & 32.6   & 58.5   & 76.4 & 88.2 & 18.5         & 50.1     \\ \bottomrule
\end{tabular}}
\label{table:naive_ensemble}
\end{table}

%% file: tables/supplement_table.tex
\begin{table}[t]
\caption{Performance of adaptive reasoning models.}
\label{table:sft_adaptive_analysis}
\vspace{-2mm}
\resizebox{0.98\textwidth}{!}{
\begin{tabular}{l|cccc|cccc|c}
\toprule
\multirow{2}{*}{\textbf{Model}} & \multicolumn{4}{c}{\textbf{Math-Oriented}}  & \multicolumn{4}{|c|}{\textbf{General Scenarios}}          & \multirow{2}{*}{\textbf{\begin{tabular}[c]{@{}c@{}}Average\\ Acc.\end{tabular}}} \\ \cmidrule{2-9}
                                & MathVista & MathVision & MathVerse & WeMath & MMStar & V*   & POPE  & SpaScore &                                                                                  \\ \midrule
Qwen2.5-VL-3B                   & 62.3      & 21.2       & 31.2      & 21.1   & 53.1      & 73.5 & 86.1  & 16.2         & 45.6                                                                             \\ \midrule
\multicolumn{10}{c}{Stage 1 SFT-based Models} \\ \midrule
AdaVaR-SFT-3B                   & 60.4      & 23.2       & 33.0      & 29.5   & 54.7     & 75.5 & 89.1  & 16.8         & 47.8                                                                             \\
GRD\%                           & 31\%      & 5\%        & 1\%       & 78\%   & 90\%    & 98\% & 94\%  & 96\%         & -                                                                                \\ \midrule
\multicolumn{10}{c}{Stage 2 RL-based Models} \\ \midrule

AdaVaR-3B                       & 69.8      & 24.5       & 35.2      & 33.8   & 59.3    & 77.1   & 88.2  & 18.9         & 50.8                                                                             \\
GRD\%                           & 2\%      & 0\%      & 0\%     & 0\%  & 51\%     & 99\% & 100\% & 59\%       & -                                                                                \\ \midrule
\multicolumn{10}{c}{SFT-Guided Adaptive Reasoning Model} \\ \midrule
SFT-Adaptive-3B                      &  59.1     &  21.2     & 30.7      & 29.1     &  52.8        & 77.5   & 89.5  &  17.6        & 47.2  \\
GRD\% & 86\%   & 90\%  & 59\%  & 100\%   & 100\%   & 100\%  & 100\%   & 100\%   & -
\\ \bottomrule
\end{tabular}}
\vspace{-2mm}
\end{table}

%% file: tables/response_type.tex
\begin{table}[t]
\caption{Proportions of different response types from AdaVaR-3B across different datasets.}
\vspace{-2mm}
\resizebox{0.98\textwidth}{!}{
\begin{tabular}{l|cccccccc}
\toprule
Response Type   & MathVista & MathVision & MathVerse & WeMath  & MMStar  & VStar   & POPE    & SpaScore \\ \midrule
Correct         & 69.80\%   & 24.50\%    & 35.20\%   & 33.80\% & 59.30\% & 77.10\% & 88.20\% & 18.90\%      \\
Selection Error & 4.80\%    & 6.30\%     & 7.20\%    & 9.10\%  & 8.80\%  & 2.60\%  & 2.90\%  & 10.50\%      \\
Reasoning Error & 25.40\%   & 69.20\%    & 57.60\%   & 57.10\% & 31.90\% & 20.30\% & 8.90\%  & 70.60\%      \\ \midrule
SE/(SE+RE)      & 15.89\%   & 8.34\%     & 11.11\%   & 13.75\% & 21.62\% & 11.35\% & 24.58\% & 12.95\%      \\ \bottomrule
\end{tabular}}
\label{table:response_type}
\end{table}

%% file: tables/error2question_type.tex
\begin{table}[t]
\caption{Distribution of selection errors across different question types and selected modes.}
\vspace{-2mm}
\centering
\resizebox{0.8\textwidth}{!}{
\begin{tabular}{l|ccccc|c}
\toprule
Selected Mode & Science & Perception & Math & Logical Reasoning & Instance Reasoning & Total \\ \midrule
GRD Mode      & 3       & 43         & 0    & 12                & 10                 & 68    \\
TXT Mode      & 33      & 0          & 14   & 12                & 12                 & 71    \\ \midrule
Total         & 36      & 43         & 14   & 24                & 22                 & 139   \\ \bottomrule
\end{tabular}}
\label{table:error2question_type}
\end{table}

%% file: tables/error2image_type.tex
\begin{table}[t]
\caption{Distribution of selection errors across different image types and selected modes.}
\vspace{-2mm}
\centering
\resizebox{0.8\textwidth}{!}{
\begin{tabular}{l|cccccc|c}
\toprule
Selected Mode & Natural Images & Diagram & Art & Geometry & Document & Others & Total \\ \midrule
GRD Mode      & 59             & 4       & 5   & 0        & 0        & 0      & 68    \\
TXT Mode      & 8              & 46      & 6   & 6        & 4        & 1      & 71    \\ \midrule
Total         & 67             & 50      & 11  & 6        & 4        & 1      & 139   \\ \bottomrule
\end{tabular}}
\label{table:error2image_type}
\end{table}

%% file: tables/train_efficiency.tex
\begin{table}[t]
\caption{Comparison between the training efficiency of GRPO and AdaGRPO. Theoretical FLOPs are estimated within a single optimization step: based on Qwen2.5-VL-3B, using an image of size 1024×768, a question of 260 tokens, and KV cache enabled. Empirical wall-clock time is measured by running 100 steps during the training process and averaging the results.}
\vspace{-2mm}
\resizebox{0.98\textwidth}{!}{
\begin{tabular}{l|ccccc|ccccc}
\toprule
\multirow{2}{*}{Algorithm} & \multicolumn{5}{c|}{Theoretical FLOPs}                                                                                                                                                                                                                                                    & \multicolumn{5}{c}{Empirical Wall-lock Time (seconds)}                                                                                                                                                                                                                                   \\ \cmidrule{2-11} 
                           & \begin{tabular}[c]{@{}c@{}}Rollout\\ Generation\end{tabular} & \begin{tabular}[c]{@{}c@{}}Reward\\ Calculation\end{tabular} & \begin{tabular}[c]{@{}c@{}}Advantage\\ Calculation\end{tabular} & \multicolumn{1}{c|}{\begin{tabular}[c]{@{}c@{}}Objective\\ Forward\end{tabular}} & Total  & \begin{tabular}[c]{@{}c@{}}Rollout\\ Generation\end{tabular} & \begin{tabular}[c]{@{}c@{}}Reward\\ Calculation\end{tabular} & \begin{tabular}[c]{@{}c@{}}Advantage\\ Calculation\end{tabular} & \multicolumn{1}{c|}{\begin{tabular}[c]{@{}c@{}}Objective\\ Forward\end{tabular}} & Total \\ \midrule
AdaGRPO                    & 20.31T                                                       & 0                                                            & 0                                                               & \multicolumn{1}{c|}{111.9T}                                                      & 132.2T & 18.45                                                        & 1.40E-02                                                     & 2.20E-04                                                        & \multicolumn{1}{c|}{1.12}                                                        & 19.58 \\
GRPO                       & 20.47T                                                       & 0                                                            & 0                                                               & \multicolumn{1}{c|}{113.9T}                                                      & 134.4T & 19.18                                                        & 1.20E-02                                                     & 1.10E-03                                                        & \multicolumn{1}{c|}{1.17}                                                        & 20.36 \\ \bottomrule
\end{tabular}}
\label{table:training_efficiency}
\end{table}

%% file: tables/inference_efficiency.tex
\begin{table}[t]
\caption{Inference time (seconds) per sample across different models and datasets, measured with 8 A100 GPUs and accelerated by the vLLM framework~\citep{kwon2023efficient}.}
\vspace{-2mm}
\resizebox{0.95\textwidth}{!}{
\begin{tabular}{l|cccccccc|c}
\toprule
Model           & MathVista & MathVision & MathVerse & WeMath & MMStar & POPE & V*   & SpaScore & Avg. \\ \midrule
AdaVaR-3B       & 0.32      & 0.95       & 0.45      & 0.48   & 0.18   & 0.07 & 0.25 & 0.87     & 0.36 \\
Text-SFT-RL     & 0.32      & 0.91       & 0.44      & 0.53   & 0.21   & 0.11 & 0.23 & 0.74     & 0.37 \\
Grounded-SFT-RL & 0.31      & 0.85       & 0.39      & 0.4    & 0.25   & 0.07 & 0.27 & 0.84     & 0.34 \\ \midrule
\# samples      & 1000      & 3040       & 3940      & 1740   & 1500   & 9000 & 191  & 1400     & -    \\ \bottomrule
\end{tabular}}
\label{table:inference_efficiency}
\end{table}

%% file: tables/sft_data_ratio.tex
\begin{table}[t]
\caption{The impact of data proportions among different modes during the SFT stage. Each cell is presented in the format `accuracy / proportion of selecting the grounded mode'.}
\vspace{-2mm}
\resizebox{\textwidth}{!}{
\begin{tabular}{lcccccccccc}
\toprule
Model         & \multicolumn{1}{c|}{SFT Ratio}   & MathVista    & MathVision  & MathVerse  & WeMath      & MMStar      & V*           & POPE         & \multicolumn{1}{c|}{SpatialScore} & Avg. Acc \\ \midrule
Qwen2.5-VL-3B & \multicolumn{1}{c|}{-}           & 62.3         & 21.2        & 31.2       & 21.1        & 53.1        & 73.5         & 86.1         & \multicolumn{1}{c|}{16.2}         & 45.6     \\ \midrule
\multicolumn{11}{c}{SFT-based Models}                                                                                                                                                               \\ \midrule
AdaVaR-SFT-3B & \multicolumn{1}{c|}{GRD:TXT=1:1} & 60.4 / 31\%  & 23.2 / 5\%  & 33.0 / 1\% & 29.5 / 78\% & 54.7 / 90\% & 75.5 / 98\%  & 89.1 / 94\%  & \multicolumn{1}{c|}{16.8 / 96\%}  & 47.8     \\
AdaVaR-SFT-3B & \multicolumn{1}{c|}{GRD:TXT=3:1} & 57.1 / 52 \% & 21.5 / 10\% & 31.2 / 3\% & 29.4 / 84\% & 51.4 / 96\% & 75.3 / 100\% & 89.6 / 100\% & \multicolumn{1}{c|}{17.0 / 100\%} & 46.6     \\
AdaVaR-SFT-3B & \multicolumn{1}{c|}{GRD:TXT=1:3} & 61.0 / 21\%  & 23.1 / 3\%  & 32.5 / 0\% & 30.0 / 40\% & 55.0 / 75\% & 75.0 / 95\%  & 86.1 / 81\%  & \multicolumn{1}{c|}{16.5 / 92\%}  & 47.2     \\ \midrule
\multicolumn{11}{c}{RL-based Models}                                                                                                                                                                \\ \midrule
AdaVaR-3B     & \multicolumn{1}{c|}{GRD:TXT=1:1} & 69.8 / 2\%   & 24.5 / 0\%  & 35.2 / 0\% & 33.8 / 0\%  & 59.3 / 51\% & 77.1 / 99\%  & 88.2 / 100\% & \multicolumn{1}{c|}{18.9 / 59\%}  & 50.8     \\
AdaVaR-3B     & \multicolumn{1}{c|}{GRD:TXT=3:1} & 66.8 / 6\%   & 24.1 / 1\%  & 34.8 / 0\% & 32.4 / 2\%  & 57.7 / 64\% & 76.8 / 100\% & 88.9 / 100\% & \multicolumn{1}{c|}{17.4 / 80\%}  & 49.9     \\
AdaVaR-3B     & \multicolumn{1}{c|}{GRD:TXT=1:3} & 68.8 / 1\%   & 24.8 / 0\%  & 35.2 / 0\% & 34.3 / 0\%  & 58.1 / 43\% & 76.4 / 99\%  & 87.2 / 100\% & \multicolumn{1}{c|}{16.6 / 40\%}  & 50.2     \\ \bottomrule
\end{tabular}}
\label{table:ablation_sft_data_ratio}
\end{table}

%% file: tables/mode_prefix_ablation.tex
\begin{table}[t]
\caption{The impact of mode prefix formatting. For AdaVaR models, each cell is presented in the format `accuracy / proportion of selecting the grounded mode'.}
\vspace{-2mm}
\resizebox{\textwidth}{!}{
\begin{tabular}{cc|cccccccc|c}
\toprule
Model         & Mode Prefix                                                     & MathVista   & MathVision & MathVerse  & WeMath     & MMStar      & V*           & POPE         & SpatialScore & Avg. Acc \\ \midrule
Qwen2.5-VL-3B & -                                                               & 62.3        & 21.2       & 31.2       & 21.1       & 53.1        & 73.5         & 86.2         & 16.2         & 45.6     \\
AdaVaR-3B     & \textless{}text\textgreater{}, \textless{}ground\textgreater{}  & 69.8 / 2\%  & 24.5 / 0\% & 35.2 / 0\% & 33.8 / 0\% & 59.3 / 51\% & 77.1 / 99\%  & 88.2 / 100\% & 18.9 / 59\%  & 50.8     \\
AdaVaR-3B     & \textless{}mode1\textgreater{}, \textless{}mode2\textgreater{} & 69.6 / 2 \% & 23.8 / 0\% & 35.0 / 0\% & 33.5 / 0\% & 59.5 / 45\% & 77.5 / 100\% & 89.0 / 100\% & 17.6 / 55\%  & 50.7     \\ \bottomrule
\end{tabular}}
\label{table:ablation_on_mode_prefix}
\vspace{-2mm}
\end{table}

%% file: tables/sft_ablation.tex
\begin{table}[t]
\caption{Ablation study on the SFT stage.}
\vspace{-2mm}
\centering
\resizebox{0.9\textwidth}{!}{
\begin{tabular}{l|cccccccc|c}
\toprule
Model         & MathVista & MathVision & MathVerse & WeMath & MMStar & V*    & POPE  & SpaScore & Avg. Acc \\ \midrule
Qwen2.5-VL-3B & 62.3      & 21.2       & 31.2      & 21.1   & 53.13  & 73.52 & 86.16 & 16.24    & 45.6     \\
AdaVaR-3B     & 69.8      & 24.5       & 35.2      & 33.8   & 59.3   & 77.1  & 88.2  & 18.9     & 50.8     \\
w/o SFT       & 63.2      & 21.7       & 31.3      & 26.9   & 54.1   & 73.2  & 82.4  & 18.6     & 46.5     \\ \bottomrule
\end{tabular}}
\label{table:ablation_sft_stage}
\end{table}

%% file: tables/system_prompt.tex
\begin{table}[t]
\caption{The impact of different system prompts. Each cell is presented in the format `accuracy / proportion of selecting the grounded mode'.}
\vspace{-2mm}
\resizebox{\textwidth}{!}{
\begin{tabular}{lcc|cccccccc|c}
\toprule
Model           & train prompt & eval prompt & MathVista   & MathVision & MathVerse  & WeMath     & MMStar      & V*           & POPE         & SpaScore    & Avg. Acc \\ \midrule
Qwen2.5-VL-3B   & -            & -           & 62.3        & 21.2       & 31.2       & 21.1       & 53.1        & 73.5         & 86.2         & 16.2        & 45.6     \\
AdaVaR-3B       & full         & full        & 69.8 / 2\%  & 24.5 / 0\% & 35.2 / 0\% & 33.8 / 0\% & 59.3 / 51\% & 77.1 / 99\%  & 88.2 / 100\% & 18.9 / 59\% & 50.8     \\
AdaVaR-3B       & full         & think-only  & 69.1 / 3 \% & 24.3 / 0\% & 36.1 / 0\% & 33.0 / 1\% & 57.8 / 59\% & 76.4 / 100\% & 88.6 / 100\% & 18.6 / 62\% & 50.5     \\
w/o full prompt & think-only   & think-only  & 69.5 / 2\%  & 25.1 / 0\% & 36.1 / 0\% & 33.2 / 0\% & 58.4 / 52\% & 77.0 / 100\% & 88.3 / 100\% & 18.5 / 59\% & 50.8     \\ \bottomrule
\end{tabular}}
\label{table:system_prompt}
\end{table}

%% file: tables/data_order.tex
\begin{table}[t]
\caption{The impact of data order in curiculum learning.}
\vspace{-2mm}
\centering
\resizebox{0.98\textwidth}{!}{
\begin{tabular}{lc|cccccccc|c}
\toprule
Model       & Data Order   & MathVista & MathVision & MathVerse & WeMath & MMStar & V*   & POPE & SpaScore & Avg. Acc \\ \midrule
AdaVaR-3B   & easy-to-hard & 69.8      & 24.5       & 35.2      & 33.8   & 59.3   & 77.1 & 88.2 & 18.9     & 50.8     \\
AdaVaR-3B-R & hard-to-easy & 68.1      & 24.2       & 35.2      & 32.3   & 58.5   & 76.4 & 87.3 & 18.1     & 50.0     \\ \bottomrule
\end{tabular}}
\label{table:ablation_on_data_order}
\end{table}

%% file: tables/gaussian_blur.tex
\begin{table}[t]
\caption{Robustness against image noise. For AdaVaR, each cell is presented in the format `accuracy / proportion of selecting the grounded mode'.}
\vspace{-2mm}
\centering
\resizebox{\textwidth}{!}{
\begin{tabular}{lc|cccccccc|c}
\toprule
Model         & Gaussian Blur & MathVista  & MathVision & MathVerse  & WeMath     & MMStar      & V*          & POPE         & SpaScore    & Avg. Acc \\ \midrule
Qwen2.5-VL-3B & No            & 62.3       & 21.2       & 31.2       & 21.1       & 53.1        & 73.5        & 86.1         & 16.2        & 45.6     \\
AdaVaR-3B     & No            & 69.8 / 2\% & 24.5 / 0\% & 35.2 / 0\% & 33.8 / 0\% & 59.3 / 51\% & 77.1 / 99\% & 88.2 / 100\% & 18.9 / 59\% & 50.8     \\
AdaVaR-3B     & Yes           & 68.8 / 2\% & 24.3 / 0\% & 35.1 / 0\% & 33.7 / 0\% & 58.3 / 45\% & 75.8 / 99\% & 88.2 / 98\%  & 18.5 / 54\% & 50.5     \\
Text-SFT-RL   & No            & 67.2       & 23.52      & 36.1       & 32.19      & 56          & 70.7        & 85.6         & 18.8        & 48.8     \\
GRD-SFT-RL    & No            & 62.6       & 22.17      & 34.4       & 33         & 55.4        & 77          & 87.4         & 18.4        & 48.7     \\ \bottomrule
\end{tabular}}
\label{table:image_blur}
\end{table}